\documentclass{article}
\usepackage{ijcai24}

\usepackage{times}
\usepackage{soul}
\usepackage{url}
\usepackage[hidelinks]{hyperref}
\usepackage[utf8]{inputenc}
\usepackage[small]{caption}
\usepackage{graphicx}
\usepackage{amsmath}
\usepackage{amsthm}
\usepackage{booktabs}
\usepackage{algorithm}
\usepackage{algorithmic}
\usepackage[switch]{lineno}

\usepackage{amssymb}
\usepackage{bm}
\usepackage{tcolorbox}
\usepackage{tikz}
\usepackage{listings}
\usepackage{xcolor}

\usetikzlibrary{calc}

\newtheorem{example}{Example}

\title{HypBO: Accelerating Black-Box Scientific Experiments Using Experts’ Hypotheses}

\author {
    Abdoulatif Cissé$^{1,3}$\and
    Xenophon Evangelopoulos$^{1,3}$\and
    Sam Carruthers$^{1,3}$\and
    Vladimir V. Gusev$^2$\and
    Andrew I. Cooper$^{1,3}$\\
\affiliations
$^1$Department of Chemistry, University of Liverpool, England, UK\\
$^2$Department of Computer Science, University of Liverpool, England, UK\\
$^3$Leverhulme Research Centre for Functional Materials Design, University of Liverpool, England, UK\\
\emails
\{abdoulatif.cisse, evangx, sgscarru, vladimir.gusev, aicooper\}@liverpool.ac.uk
}

\begin{document}

\maketitle

\begin{abstract}
Robotics and automation offer massive accelerations for solving intractable, multivariate scientific problems such as materials discovery, but the available search spaces can be dauntingly large. Bayesian optimization (BO) has emerged as a popular sample-efficient optimization engine, thriving in tasks where no analytic form of the target function/property is known. Here, we exploit expert human knowledge in the form of hypotheses to direct Bayesian searches more quickly to promising regions of chemical space. Previous methods have used underlying distributions derived from existing experimental measurements, which is unfeasible for new, unexplored scientific tasks. Also, such distributions cannot capture intricate hypotheses. Our proposed method, which we call HypBO, uses expert human hypotheses to generate improved seed samples. Unpromising seeds are automatically discounted, while promising seeds are used to augment the surrogate model data, thus achieving better-informed sampling. This process continues in a global versus local search fashion, organized in a bilevel optimization framework. We validate the performance of our method on a range of synthetic functions and demonstrate its practical utility on a real chemical design task where the use of expert hypotheses accelerates the search performance significantly.
\end{abstract}

\section{Introduction}
\label{sec:intro}
Bayesian Optimization (BO) is a valuable tool for optimizing experiments in chemistry and materials science, where experiments are costly and time-consuming ~\cite{c:Taking_the_human_out_of_the_loop}. Experimental design methods often involve exhaustive exploration of the parameter space. By contrast, BO offers an efficient framework leveraging Bayesian inference to guide the iterative exploration of the space, ultimately maximizing the target experiment property ~\cite{c:EGO}.

Formally, BO aims to find the global optimum in the following problem: 
\begin{equation}
    \small
    x^{*} = \underset{x \in \mathcal{X}}{argmax}\ f(x), \label{BO}
\end{equation}
where \(f: \mathcal{X} \rightarrow \mathbb{R}\) is a continuous function over the \(d\)-dimensional input space \(\mathcal{X} \in \mathbb{R}^{d}\). Generally, the underlying analytical form of $f(\cdot)$ is unknown, making it a black-box function. The core principle of BO lies in the construction of a probabilistic model, typically a Gaussian Process (GP) ~\cite{c:gaussian_processes}, which serves as a \emph{surrogate model} for $f(\cdot)$. This surrogate model is updated iteratively as new experimental data become available, allowing for the refinement of target predictions. The model's uncertainty is quantified, and an \emph{acquisition function} is employed to select the next set of experimental parameters to evaluate, balancing exploration (sampling in unexplored regions) and exploitation (focusing on promising regions).

Injecting domain-specific knowledge into BO to boost optimization performance has gained significant recent attention, especially for scientific tasks ~\cite{c:Incorporating_Expert_Prior_in_BO_via_Space_Warping}, aiming to alleviate the resource-intensive surrogate model construction. In particular, recent studies have used expert knowledge as user-specified priors over possible optima to guide the search toward promising regions ~\cite{c:piBO,c:Incorporating_Expert_Prior_Knowledge}. While this has shown promising performance in various tasks, it is difficult in many scientific problems to realize external knowledge in the form of a prior distribution. Furthermore, the optimization landscapes of such problems often resemble a needle-in-a-haystack manifold ~\cite{c:Zombi}, and inaccurate prior knowledge distributions can introduce negative bias in the problem and quickly degrade performance. More recently, human-in-the-loop (HIL) approaches have emerged where an interactive optimization framework enables experts to implicitly add knowledge to the problem in the form of feedback on the quality of the samples within the experimental loop ~\cite{c:BO_Augmented_with_Actively_Elicited_Expert_Knowledge}. However, this knowledge is implicit and sample-specific and can often lead to local optima entrapment. Another category of methods introduces domain-specific knowledge in the form of hard constraints in the problem ~\cite{c:hernandez2015constraints}, which can, however, over-restrict the search in practice.

\begin{figure}[t]
\centering
\includegraphics[width=0.4\columnwidth]{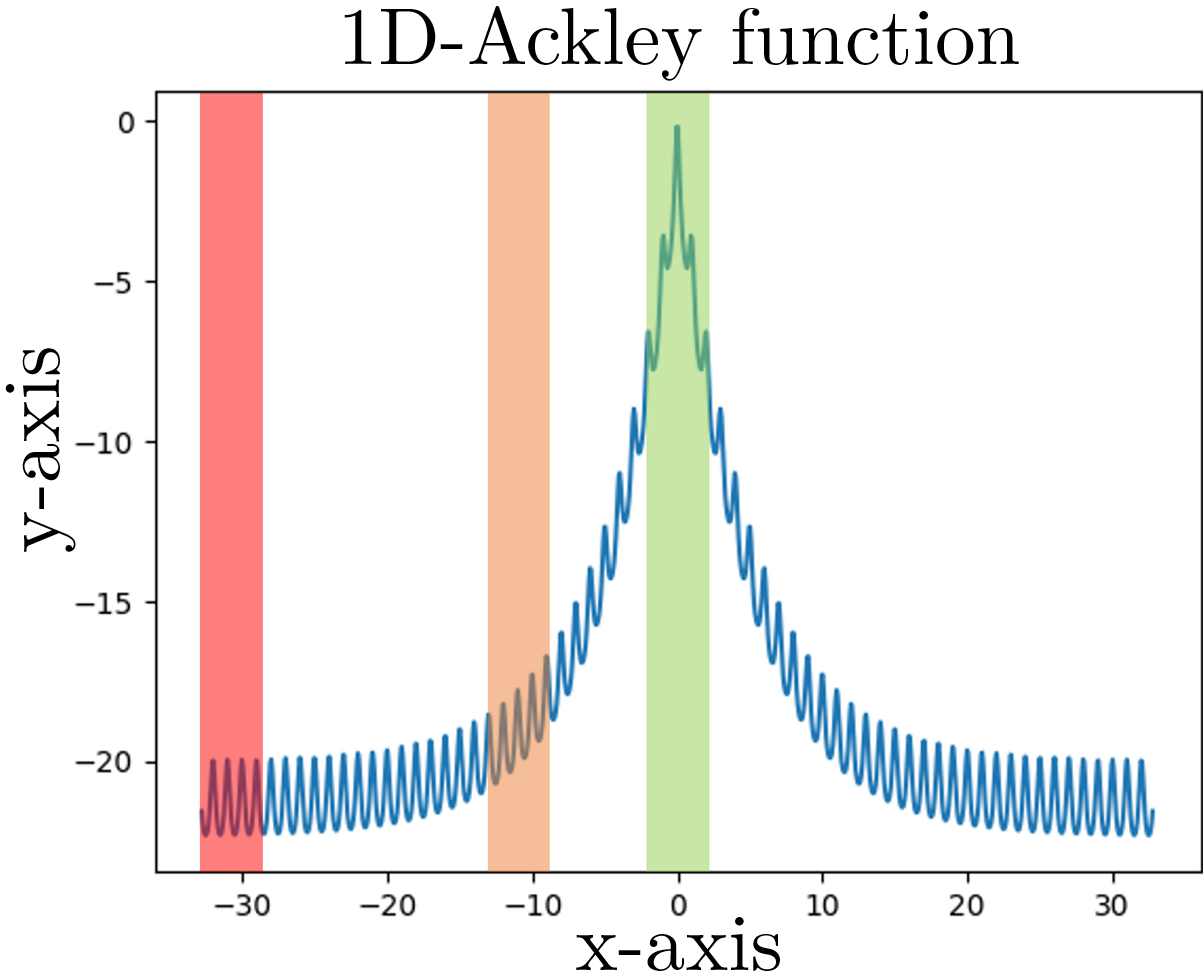}
\caption{Illustration of three hypothesis locations in the form of confidence regions on the 1D Ackley function. The different colors (red, orange, and green) correspond to different levels of confidence (poor, weak, and good, respectively).}
\label{fig:hypotheses_location}
\end{figure}

In this paper, we propose a novel approach to inject domain knowledge using input from domain experts to direct the search to more fruitful regions. We specifically represent domain knowledge as human hypotheses or conjectures that are realized as intervals of confidence, i.e., constraints on the parameter space. Figure~\ref{fig:hypotheses_location} demonstrates three representative hypothesis regions within the input space of a one-dimensional Ackley function where the input region around zero is clearly the most promising hypothesis. The various hypotheses are realized as local Gaussian processes (GPs) restricted to the constrained space, and their utility is being iteratively evaluated by a global GP, which in turn expands or shrinks the global search space accordingly. Our approach treats the human hypotheses as soft constraints and hence avoids over-restricting the search or getting stuck in local optima. We formulate our approach in a bilevel optimization framework where the lower level evaluates the various hypotheses, and the upper level integrates the useful ones in the search. The methodology is further detailed in Section~\ref{sec:methodology}. 

We test the proposed methodology on a materials design simulation where a set of different chemical hypotheses are injected to guide the search to more fruitful solutions faster. We show that hypotheses with favorable conditions accelerate the search and also improve performance. Interestingly, unfavorable hypotheses do not appear to bias the search negatively in the long run. Extensive synthetic tests further demonstrate that our method maintains a competitive and robust performance overall.

The remainder of this paper is organized as follows. Section  \ref{sec:related_works} presents recent works about expert knowledge integration in BO, while Section \ref{sec:methodology} describes the proposed methodology. The robustness and performance of our algorithm are evaluated and discussed in Section \ref{sec:experiments}. Finally, Section \ref{sec:conclusions} summarizes our work and introduces future directions.

\section{Related Works}
\label{sec:related_works}
Knowledge distillation has recently been at the center of attention in the BO literature to address issues such as the ``cold'' start problem where the initial points, usually selected randomly, fail to adequately capture the optimization objective's landscape. Transfer learning~\cite{c:transfer_learning} has been widely used to extract and use knowledge from previous BO executions to aid in warming up and enhancing optimization~\cite{c:transfer_learning_BO}. Furthermore, it has also been used effectively in chemical reaction optimization ~\cite{c:transfer_learning_in_chemical_reaction_opti} to bias the search space by weighting the current acquisition function with past predictions. 

Another approach to improving BO through the incorporation of domain knowledge involves the use of similarities between points in the search space. Gryffin~\cite{c:gryffin} uses user-provided physicochemical descriptors to navigate the search space more efficiently by identifying similarities between individual options based on those descriptors. However, when using a large number of descriptors, spurious correlations can occur between descriptors and the optimized objective, leading to irrelevant descriptors being considered important. \cite{c:initial_sampling_clustering} suggest using a clustering-based initial sample selection method for optimizing chemical reaction conditions with BO based on a high correlation between molecular descriptors and clustering in chemical space. However, as clustering is based on unsupervised learning, there is a need for expert knowledge to connect it with experimental results; also, not all scientific problems can be codified using molecular descriptors.

Other approaches inject expert prior beliefs as priors to guide the optimization process. ~\cite{c:Incorporating_Expert_Prior_Knowledge} combined prior user beliefs with observed data to compute the posterior distribution via repeated Thompson sampling. This approximates new sampling points using a linear combination of posterior samplings. BOPrO ~\cite{c:BO_with_a_Prior_for_the_Optimum} uses a prior that is provided by the user and a data-driven model to generate a pseudo-posterior. Similarly, $\pi$BO ~\cite{c:piBO} generates a pseudo-posterior by integrating prior beliefs into the acquisition function as a decaying multiplicative factor to improve sampling. Both of these methods, and ColaBO ~\cite{c:ColaBO} which augments the surrogate model with a user-defined prior, are limited to one expert prior, and the use of priors cannot capture intricate knowledge. ~\cite{c:hypothesis_learning} co-navigates a hypothesis space and the experimental space through a hypothesis learning approach that combines multiple hypotheses as probabilistic models with reinforcement learning. However, major drawbacks of this approach are the difficulty of representing a hypothesis into a probabilistic model from the functional form of the black-box model, the computational cost and time of applying Bayesian Inference for each model, and the assumption that only one out of the hypothesis pool is the correct one.

Preference learning can also enrich BO through domain knowledge. \cite{c:BO_Augmented_with_Actively_Elicited_Expert_Knowledge} obtain expert opinions by querying them with pairwise comparisons, thereby approximating the shape of the objective function. \cite{c:Human_AI_Collaborative_BO} take a slightly different approach by allowing experts to provide a pair of good and bad points, which are then used to fine-tune the BO's surrogate model by replacing its current optimal hyperparameters with ones that align more closely with the expert's cognitive model. Such approaches are promising but risk biasing the optimizer, which could mimic the user's beliefs and result in suboptimal solutions.

Various methods can restrict the search space to regions assumed by the optimizer to contain the optimum. TuRBO~\cite{c:TurBO} uses multiple independent GP surrogate models within different trust regions to conduct simultaneous BO runs, and a multi-armed bandit (MAB) strategy~\cite{c:MAB} to choose which local optimizations to continue. TREGO~\cite{c:TREGO} proposed alternating between global BO and a trust region-based policy for the local phase when the global BO is failing. Alternatively, LA-MCTS~\cite{c:LAMCTS} proposes the use of Monte Carlo tree search to learn which subregions of the search space are more likely to contain good objective values. The space is then recursively partitioned based on optimization performance. Similarly, ZoMBI~\cite{c:Zombi} iteratively keeps the best sample points found so far and ``zooms'' in the sampling search bounds towards the region formed by those samples. While our approach shares similarities with the above in terms of segmenting the search space, ours sets itself apart by integrating human-friendly hypothesis-based constraints that avoid over-restricting the search. This differentiation emphasizes the innovative use of expert knowledge in our approach, particularly in complex scenarios where such insights are crucial. More importantly, we propose a generalized strategy that allows the injections of multiple expert hypotheses, such as those derived from a multi-person research team, where promising seeds from those hypotheses augment BO's surrogate model data to achieve better-informed sampling.

\section{Methodology} 
\label{sec:methodology}
In this section, we propose a novel BO methodology that uses experts' background knowledge in the form of optimality hypotheses to guide search space exploration more effectively. Let $\{\mathcal{H}\}_{j=1}^J$ be a set of manually designed hypotheses w.r.t. promising areas (subspaces) formulated in hyperrectangles and explicitly specified by experts through a system of $p$ equations and $q$ inequalities describing an interval of confidence in the search space:
\begin{equation}
    \small
    \label{eq:hypotheses_(in)equalities}
    \begin{aligned}
        Ax &= b, \\ 
         Bx &\le c,
    \end{aligned}
\end{equation}
where $A\in \mathbb{R}^{p \times d}$ and $B\in \mathbb{R}^{q \times d}$ are coefficient matrices, and $b \in \mathbb{R}^p$ and $c \in \mathbb{R}^p$ are solution vectors. This system filters the space $\mathcal{X}$ and forms a solution set, which we refer to as hypothesis subspace $\mathcal{H}_j$. In scientific experiments, experts are accustomed to thinking about parameters and conditions in terms of ranges and relationships; formulating these as mathematical constraints is more intuitive than manually setting up a prior distribution. More details and examples on how to create a hypothesis are in the Supplementary Material (SM).

Our goal is to inject practitioners' expertise into the problem at hand by attending to specific regions of the search space based on domain hypotheses, minding at the same time not to over-restrict and negatively bias the search. We, therefore, model the various different hypotheses as \textit{local} GPs acting on a constrained parameter space and using their output samples as \textit{seeds} for the \textit{global} search which in turn is realized by a \textit{global} GP. The utility of the seeds can be measured using any standard acquisition function, and the top-performing seeds are selected and fed through to the global search. This iterative global-versus-local Bayesian search takes effect interchangeably and is organized in a parametric bilevel optimization framework, which in this instance can be solved sequentially as a two-stage decision problem with each level’s variables treated as a parameter for the other~\cite{c:koppe2010bilevel}. The following paragraphs detail and formalize the proposed optimization framework.

\subsection{Upper Level}
In this level, we seek to find the global maximum of $f(\cdot)$ in \eqref{BO} as in any standard BO task where an acquisition function $\alpha(\cdot)$ is maximized to obtain new candidate samples and evaluate them across its iterations
\begin{equation}
    \small
     x^{*} = \underset{x \in \mathcal{X}}{argmax}\ \alpha (x,\mathcal{D}) \label{eq:acquisition_function}
\end{equation}
with \( \mathcal{D}=\{x_i,y_i=f(x_i)\}^{n}_{i=1}\) being the observation dataset. To compute $\alpha(\cdot)$, BO relies on constructing a global surrogate model of the underlying function and greatly depends on the initial samples provided as a seed when building this. An appropriate initial sampling has been shown to significantly improve the performance of the search in practice ~\cite{c:initial_sampling_clustering}. At this level, one could use practically any variant of BO, but we have empirically observed that using the LA-MCTS algorithm ~\cite{c:LAMCTS} helps the search to focus on promising regions to avoid over-exploring.

\subsection{Lower Level}
The lower level initially uses the subspaces from the given hypotheses to perform a \textit{local} search and yield a set of best-performing seed samples $\{s\}_{t=1}^T$, with $T<J$, essentially acting as soft constraints on the target objective function $f(\cdot)$. Given that we do not have any analytical information about $f(\cdot)$, we approximate it in the hypothesis subspaces by multiple local GP models $\phi_{j}\sim \mathcal{N}(\mu_{j}(x),k_{j}(x,x'))$ simultaneously, one for each hypothesis subspace $\mathcal{H}_j$. The local models $\phi_{j}$ are chosen as GP surrogate models for their robustness to noise and uncertainty ~\cite{c:gaussian_processes}. The local search is realized in a MAB fashion where getting a seed $s_t$ translates into selecting the most promising hypothesis via an implicit policy where the hypotheses are the arms before doing a local BO in that hypothesis region. This allows us to evaluate the hypotheses and steer the sampling toward promising regions.  

In the initialization phase, before the optimization loop starts, we ensure the hypothesis regions are covered by a specific strategy. For each hypothesis subspace $\mathcal{H}_j$, one random sample is drawn, which provides more informative seeds for the global search. If the number of desired initial samples ($n$) is not exhausted after allocating one sample per hypothesis, the additional random points are drawn from the entire search space to enhance diversity and exploration. In the event that the number of hypotheses exceeds $n$, we ensure that at least one additional random point is taken from the whole search space, making $m = \max(1, n-J)$. This strategy guarantees that each hypothesis is represented from the dataset and prevents scenarios where no initial points fall within the hypothesis regions.

As the optimization progresses, the hypothesis subspaces $\mathcal{H}_j$ will potentially have more samples, which will update the local models, better evaluating the hypotheses and producing better seeds. Stopping criteria for each level and global convergence are discussed below in Section~\ref{sec:convergence}.

The complete bilevel framework is formalized as follows
\begin{equation} \label{eq:upper_level}
    \small
    \begin{aligned} 
        x^{*} &= \underset{x \in \mathcal{X}}{argmax}\ \alpha (x,\{(s_t, f(s_t))\}_{t=1}^T\cup\mathcal{D})\\
        \text{s.t}&\\
    \end{aligned}
    \tag{Upper}
\end{equation}

\begin{equation} \label{eq:lower_level}
    \small
    \begin{aligned} 
        & \{s\}_{t=1}^T \in \underset{x \in \bigcup_{j=1}^J \mathcal{H}_j}{argmax} \{\underset{x \in \mathcal{H}_j}{max~} \phi_{j} \}.
    \end{aligned}
    \tag{Lower}
\end{equation}

\subsection{Convergence Criteria}
\label{sec:convergence}
To maximize the information gained from good hypotheses (true), we allow the lower level to produce more seed samples until it plateaus. That is, the lower level returns seed samples until these fail to improve upon the best target value:
\begin{equation}
    \begin{gathered}
    \label{eq:lower_level_failure_limit}
    \small
    (1+\gamma) y_{\text{max}} \ge f(s_{i}) \text{ if } y_{max} \ge 0 \\
    \text{or} \\
    (1-\gamma) y_{max} \ge f(s_{i}) \text{ if } y_{max} < 0 \\
    \text{for }i=k+1,\ldots,k+l_{max}
  \end{gathered}
\end{equation}

where $y_{max}$ is the best value found, $i$ is the current iteration number, $k$ is the iteration number from which the plateauing started, \(\gamma\ \in \mathbb{R}^+\) is the growth step size, and \(l_{max} \in \mathbb{N}^+\) dictates after how many consecutive iterations we deem the lower level plateauing.\newline
To mitigate weak and poor hypotheses (false), we allow the upper level to carry the optimization from the given seeds until it plateaus. It keeps maximizing \(\alpha\) until it fails to improve upon the best target value, that is:
\begin{equation}
    \begin{gathered}
    \label{eq:upper_level_failure_limit}
    \small
    (1+\gamma) y_{\text{max}} \ge f(x_{i}) \text{ if } y_{max} \ge 0 \\
    \text{or} \\
    (1-\gamma) y_{max} \ge f(x_{i}) \text{ if } y_{max} < 0 \\
    \text{for }i=k+1,\ldots,k+u_{max}
  \end{gathered}
\end{equation}
where $u_{max} \in \mathbb{N}^+$ dictates after how many consecutive iterations we deem the upper level failed. We set $l_{max} \ll u_{max}$ to direct the search toward the hypotheses' regions if they are helping to improve while still giving the upper level the time to explore the entire search space \(\mathcal{X}\).

$\gamma$ sets the percentage of improvement over the best $y$ found so far, which is considered ``significant" for the optimization process. A larger $\gamma$ means that the algorithm requires a larger improvement to consider the level optimization as still progressing. Conversely, a smaller $\gamma$ makes the criterion for progress more strict, as even minor improvements will be considered significant. 

The optimization steps are detailed in Algorithm~\ref{alg:algorithm}.

\begin{algorithm}[!htb]
    \caption{Hypothesis Bayesian Optimization (HypBO)}
    \label{alg:algorithm}
    \textbf{Input}: Hypotheses $\{\mathcal{H}_j\}_{j=1}^J$, Number of initial samples $n$, Maximum iteration number $i_{max}$, Improvement growth size $\gamma$, Number of locally optimal samples $T$ to keep, convergence parameters $l_{max}$ and $u_{max}$
    \textbf{Output}: {\raggedright $y_{max}$ \par}
    \begin{algorithmic}[1] 
        \STATE Initialize the dataset $\mathcal{D} = \{\}$;
        \FOR{each hypothesis subspace $\mathcal{H}_j$}
            \STATE Sample $x_j$ randomly from $\mathcal{H}_j$;
            \STATE Evaluate $y_j = f(x_j)$ and add $(x_j, y_j)$ to $\mathcal{D}$;
        \ENDFOR
        \STATE Randomly sample $m = \max(1, n-J)$ points from the entire search space $\mathcal{X}$, evaluate and add them to $\mathcal{D}$;
        \STATE Set $y_{max}$ as the maximum $y$ value in $\mathcal{D}$ and $i=0$;
        
        \WHILE{$i < i_{max}$}
            \begin{tcolorbox}[colback=gray!5,colframe=black,boxsep=0pt,boxrule=0.5pt, top=0pt, bottom=1pt, before skip=2pt,after skip=2pt]
                \begin{tikzpicture}[overlay, remember picture]
                    \node[rotate=90, text=black, anchor=north] at ($(current page.east)+(-9.5 cm, 3)$) {\textbf{Lower Level}};
                \end{tikzpicture}
                    
                \STATE Set attempt without improvement count $l=0$; 
                \WHILE{$l < l_{max}$ and $i < i_{max}$} 
                    \FOR{each expert-defined hypothesis $\mathcal{H}_j$}
                        \STATE Fit a GP $ \phi_j$ within the hypothesis $\mathcal{D} \cap \mathcal{H}_{j}$;
                        \STATE Find the best sample $s_j$ maximizing $\alpha_{\phi_j}$;
                    \ENDFOR
                    \STATE Keep the best samples $\{ s_t\}_{t=1}^{T}$  w.r.t. $\alpha_{\phi_j}$;       
                    \STATE Evaluate the samples $\{ s_t\}_{t=1}^{T}$ and set $y_{t_{max}}$ as the maximum of all $y_t = f(s_t)$;
                    \STATE Increment $l$ if there is no improvement, i.e. $y_{t_{max}} \leq y_{max} + \gamma$, else reset $l$ to $0$;
                    \STATE  Update the records $\mathcal{D} \gets \mathcal{D} \cup \{ (s_t, y_t)\}_{t=1}^{T}$;
                    \STATE Update $y_{max}$ as the maximum $y$ value in $\mathcal{D}$;
                    \STATE $i\gets i+1$;
                \ENDWHILE
            \end{tcolorbox}
            
            \begin{tcolorbox}[colback=gray!5,colframe=black,boxsep=0pt,boxrule=0.5pt, top=0pt, bottom=1pt, before skip=2pt,after skip=2pt]
                \begin{tikzpicture}[overlay, remember picture]
                    \node[rotate=90, text=black, anchor=north] at ($(current page.east)+(-9.5cm, -1.7)$) {\textbf{Upper Level}};
                \end{tikzpicture}
                    
                \STATE Set attempt without improvement count $u=0$;
                \WHILE{$u < u_{max}$ and $i < i_{max}$}
                    \STATE Fit a GP on the entire search space $\mathcal{D}$;
                    \STATE Find the best sample $x^*$ maximizing $\alpha$;
                    \STATE Evaluate $x^*$, $y^*=f(x^*)$;                    
                    \STATE Increment $u$ if there is no improvement, i.e. $y^* \leq y_{max} + \gamma$, else reset $u$ to $0$;
                    \STATE Update the records $\mathcal{D} \gets \mathcal{D} \cup \{(x^*, y^*)\}$;
                    \STATE $i\gets i+1$;
                \ENDWHILE
            \end{tcolorbox}
        \ENDWHILE
        \RETURN The maximum value found, $y_{max}$
        \end{algorithmic}
\end{algorithm}

\section{Experiments}
\label{sec:experiments}
We showcase the effectiveness of our proposed method in optimizing various synthetic functions and real-world problems, such as discovering new materials. We test HypBO's performance and robustness using hypotheses of different qualities, ranging from good to poor. We also compare its performance against other BO algorithms. In Section \ref{sec:experimental_setup}, we outline the various experimental settings and comparison methods we used to benchmark our results. Sections \ref{sec:synthetic_functions} and \ref{sec:her_experiment} present the outcomes of an analytical function optimization task and a materials design problem ~\cite{c:Mobile_robotic_chemist}, respectively.

\begin{figure}[t]
    \centering
    \includegraphics[width=0.8\columnwidth]{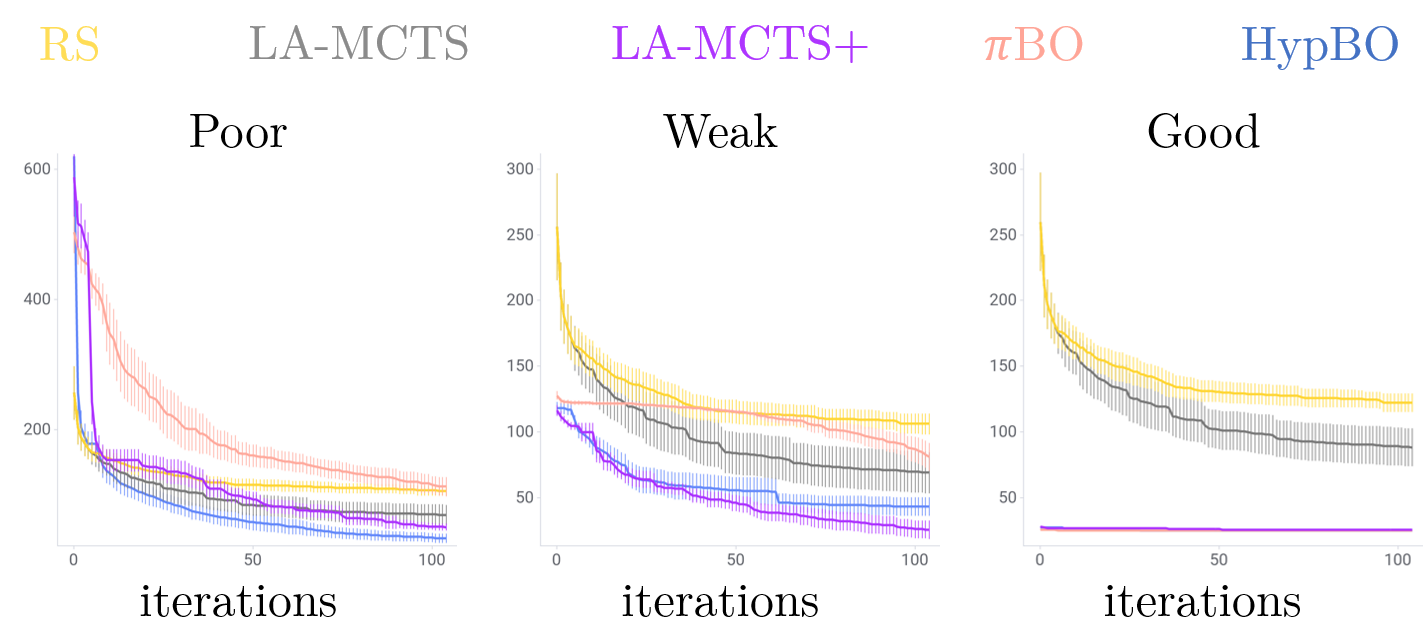}
    \caption{Comparison of RS, LA-MCTS, LA-MCTS+, $\pi$BO, and HypBO on Levy\textsubscript{d20} for various hypothesis qualities (see Figure \ref{fig:hypotheses_location}). Solid lines show the mean values, while the shaded areas represent the standard error.}
    \label{fig:single_hypothesis}
\end{figure}

\begin{figure}[t]
    \centering
    \includegraphics[width=0.95\columnwidth]{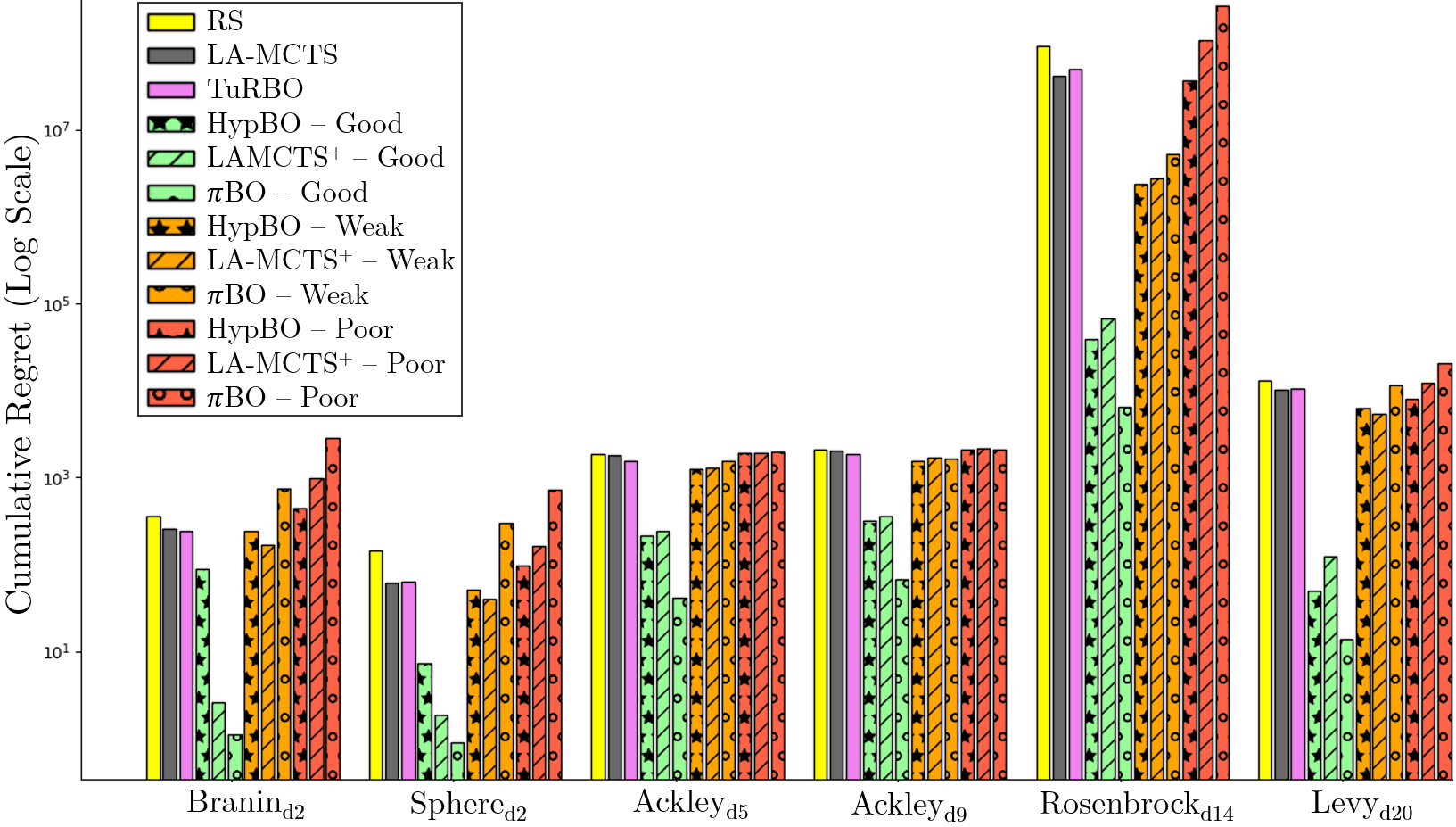}
    \caption{Cumulative regret on functions with various dimensions.}
    \label{fig:cumulative_regret_single_hypothesis.png}
\end{figure}

\subsection{Experimental Setup}
\label{sec:experimental_setup}
We evaluate HypBO's performance empirically for the following two tasks:
\begin{itemize}
    \item \textbf{Synthetic Functions} We test the precision, convergence speed, and robustness of HypBO using synthetic benchmark functions with various nonconvex landscapes and dimensionalities. The optimization performance is measured using simple and cumulative regrets, and Wilcoxon tests ~\cite{c:wilcoxon}. The maximum number of iterations is limited to 100, and the result of 50 repeated trials is reported as the mean value.
    \item \textbf{Photocatalytic Hydrogen Production} Here, we replicate the materials design problem addressed in ~\cite{c:Mobile_robotic_chemist}, aiming to maximize the hydrogen evolution rate (HER) from a mixture of different materials. We follow a more cost-effective approach and emulate that chemistry experiment by 
    interpolating new HER measurements using a GP model trained on existing experimental data points. In section \ref{sec:her_experiment}, we give a comprehensive explanation of this chemistry task. The maximum number of iterations is set to 300, and the mean value of 50 repeated trials is reported.
\end{itemize}
We further evaluate HypBO against the following baselines whose hyperparameters' values are given in the SM:
\begin{itemize}
    \item \textbf{Random Search (RS)} Random search under uniform distribution over the search space.
    \item \textbf{Trust Region Bayesian Optimization (TuRBO)} with one trust region.
    \item \textbf{Latent Action Monte Carlo Tree Search (LA-MCTS)}
    \item \textbf{LA-MCTS with hypothesis-based initial design (LA-MCTS+)} We modified the previous baseline to initialize it exclusively within the hypothesis subspaces before the regular search in the entire search space.
    \item $\bm{\pi}$\textbf{BO} This baseline uses expert knowledge throughout optimization. As described in Section \ref{sec:related_works}, it is one of the most competitive methods for priors over optimum. We converted our hypotheses into Gaussian priors centered around the hypothesis subspace center (see SM).
\end{itemize}

For all experiments, we use preset hyperparameters for HypBO. We set the lower level limit $l_{max}$ to 2, the upper level limit $u_{max}$ to 5, the number of locally optimal samples $T$ to 1, and the growth rate $\gamma$ to 0. This value of $\gamma$ essentially means that as long as any improvement is being made (no matter how small), the level optimization will continue. Note that this could be beneficial in scenarios where even small gains are valuable, but it might also make the optimization process slower or more prone to getting stuck in flat regions where minute fluctuations might appear as improvements. Ablation studies can be found in the SM. Concerning the local GP models for the hypotheses, they have a zero mean and a Matérn ($\nu=2.5$, $\lambda_i = 1$) kernel with constant scaling. Note that the kernel's hyperparameters are automatically optimized based on the experimental data to fit the models best. All experiments are warm-started with five initial points except for the photocatalyst hydrogen production experiment with mixed hypotheses, whose initial sample count is 10. Reproducibility details are available in the SM.

\subsection{Synthetic Functions}
\label{sec:synthetic_functions}
\subsubsection{Hypotheses}
A good hypothesis subspace is essentially an interval that contains the optimum, $opt$. By contrast, a weak/poor one does not. The further the weak hypothesis subspace is from the optimum, the worse it is. Here, the ``poor" hypothesis is the furthest from the optimum. The hypotheses are hyperrectangles of width $w=2$ units and centered as follows:
\begin{itemize}
    \item \textbf{Poor hypothesis} at $l_b+w/2$ where $l_b$ is the lower bound of the search space.
    \item \textbf{Weak hypothesis} at $opt-0.2*(opt-l_b)-w/2$.
    \item \textbf{Good hypothesis} at $opt$.
\end{itemize}

We assess HypBO empirically in two different settings. First, we evaluate its performance and robustness against the quality of the hypothesis. Second, we test its ability, when faced with mixed hypotheses simultaneously, to discard the weaker hypotheses and prioritize promising ones.

\subsubsection{Optimization With a Single Hypothesis}
Figure \ref{fig:single_hypothesis} shows that HypBO benefits from informative hypotheses and can also recover from weak ones. The method improves the search performance dramatically over RS, TuRBO, and LA-MCTS for a good hypothesis. The seeds from that hypothesis aid in recognizing the promising subspace and focusing efforts there, resulting in a faster location of the optimum. A similar behavior is observed in both LA-MCTS+ and $\pi$BO, whose initial sampling is entirely done in the good hypothesis region. Concerning the weak hypothesis, HypBO converges toward the optimum faster than LA-MCTS, TuRBO, and RS. In fact, the seeds coming from the weak hypothesis, by outperforming the existing samples in the dataset, direct HypBO towards the hypothesis' surroundings, too, which are more promising. As would be expected, poor hypotheses lead to a slower search in the early stages, but HypBO displays desired robustness by recovering from the poor seeds to approximately equal regret as LA-MCTS and TuRBO. However, it is interesting to note that HypBO with a poor hypothesis outperforms LA-MCTS and TuRBO in high-dimensional functions due to its more diverse initial sampling strategy. Its initial sampling strategy combines one sample from the poor hypothesis with others from across the search space, leading to a more comprehensive understanding of the overall landscape. As shown in Figure \ref{fig:cumulative_regret_single_hypothesis.png}, this approach efficacy seems to increase with the search space complexity, i.e., its dimensionality, making the advantage of diverse sampling more pronounced. Both LA-MCTS+ and $\pi$BO lag behind HypBO in these two last hypothesis scenarios as their initial sampling being entirely made of samples from the weak (respectively poor) hypothesis region is not diverse enough, and $\pi$BO's trade-off decay hyperparameter $\beta$ keeps it unnecessarily longer in that weak (respectively poor) region as shown in Figure \ref{fig:cumulative_regret_single_hypothesis.png}. This highlights HypBO's ability to exploit the explicit and implicit information the hypothesis provides faster and more intelligently. Moreover, we conducted Wilcoxon signed-rank tests ~\cite{c:wilcoxon} (at a 95\% confidence level with Bonferroni correction ~\cite{c:bonferroni}) and examined the mean and median cumulative regrets. The tests reveal that HypBO performs significantly better than RS, LA-MCTS, and TuRBO. Although the p-values for comparisons with LA-MCTS+ and $\pi$BO were not statistically significant, they were low for $\pi$BO ($p=0.06$ and $0.15$ for weak and poor hypotheses), indicating $\pi$BO's weaker performance. Moreover, HypBO showed higher median and mean regrets compared to LA-MCTS+ and $\pi$BO variants for weak and poor hypotheses, leading us to conclude that HypBO generally outperforms LA-MCTS+ and $\pi$BO.

\begin{figure}[t]
    \centering
    \includegraphics[width=0.85\columnwidth]{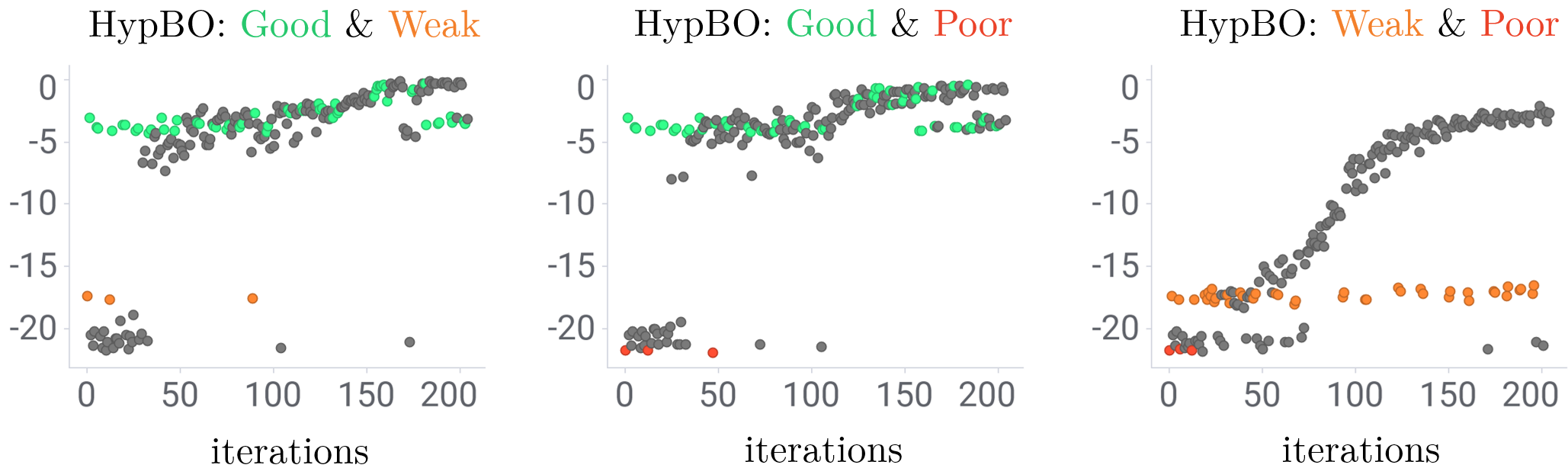}
    \caption{HypBO on the 9D-Ackley function with three different mixtures of hypotheses of various qualities for 200 iterations. Colored sample points came from the hypotheses, i.e., the lower level, while the grey ones came from the upper level.}
    \label{fig:fig_mixed_hypotheses_ackley}
\end{figure}

\subsubsection{Optimization With Mixed Hypotheses}
\begin{figure}[t]
    \centering
    \includegraphics[width=0.95\columnwidth]{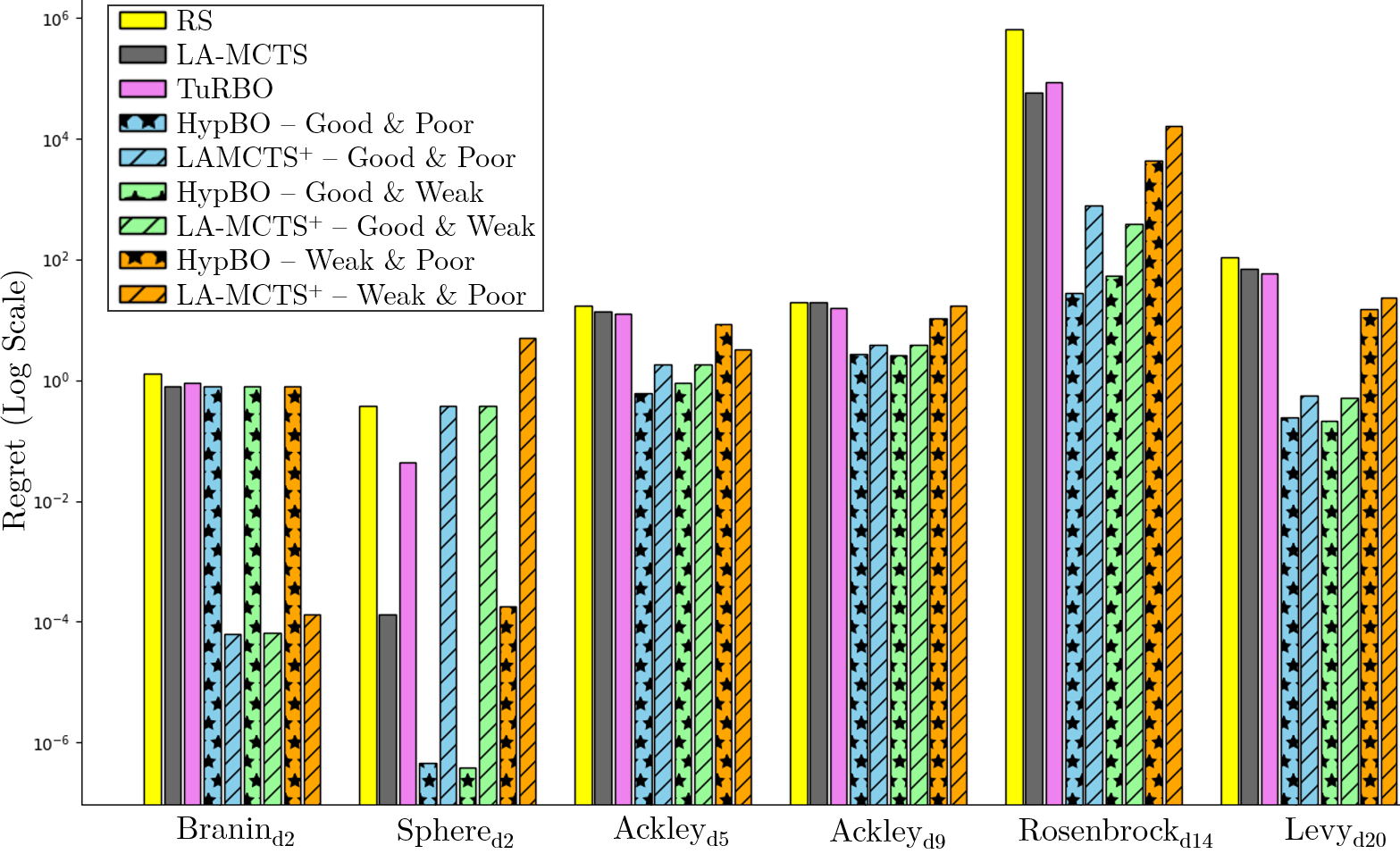}
    \caption{Regret on synthetic functions with mixed hypotheses.}
    \label{fig:fig_regret_mixed_hypotheses}
\end{figure}

We use three different binary combinations of hypotheses of varying quality to test HypBO's ability to uncover and prioritize promising hypotheses, and to discard bad ones from a pool of hypotheses. HypBO takes seeds from all the given hypotheses, which it uses to update its beliefs about each hypothesis via the MAB procedure. As the optimization progresses, it has a better representation of the hypotheses and can abandon the weaker one of the pair and select seeds from the more promising hypothesis. As shown in Figure \ref{fig:fig_mixed_hypotheses_ackley}, for the Ackley\textsubscript{d9} function, this approach allows HypBO to deselect the weaker hypothesis early on. It keeps the remaining stronger hypothesis, which it uses to expedite the search as described in the previous subsection. Figure \ref{fig:fig_regret_mixed_hypotheses} shows that these findings are consistent when applied to a variety of synthetic functions of higher dimensions. For lower dimensions, HypBO with mixed hypotheses has approximately equal regret performance to TuRBO, LA-MCTS, and LA-MCTS+ as the search space is smaller, and it becomes easier to capture the underlying behavior of the objective function. For higher dimensions, even in the case of combined weak and poor hypotheses, HypBO outperforms the other methods, demonstrating its robustness when faced with multiple items of inaccurate knowledge and the ability to use these for better sampling.

Here, the Wilcoxon tests with Bonferroni correction show no significant difference between HypBO and LA-MCTS+, albeit much lower p-values for the Good \& Poor and Good \& Weak scenarios ($p=0.15$). Along with an examination of the mean and median regrets where HypBO has lower values, we conclude that HypBO outperforms LA-MCTS+. These statistical tests also show that HypBO greatly outperformed LA-MCTS, TuRBO, and RS ($p \leq \alpha_{adjusted}$) with much lower mean and median regrets.

\subsection{Photocatalytic Hydrogen Production Optimization} \label{sec:her_experiment}
We test HypBO on a real materials design problem where we seek an optimal composition of ten materials to maximize hydrogen production via photocatalysis~\cite{c:photocatalysis}. Due to the combinatorially large search space (98,423,325 possible combinations), ~\cite{c:Mobile_robotic_chemist} used an autonomous mobile robotic chemist along with a discretized Bayesian optimizer (DBO), which can discretize the input space, to search for the optimal combination of materials.

We recast this experimental problem as a more cost-effective multivariable simulation; that is, we mapped out the chemical space by interpolating available experimental observations using a Gaussian process regression (GPR). Specifically, this GPR model has a zero mean, a Matérn ($\nu=2.5$) kernel with constant scaling and homoscedastic noise; each variable lengthscale $\lambda_i$ is initialized as its discretization step. We fitted this model against a total ``ground truth'' dataset of 1119 experimental observations supplied by the authors of~\cite{c:Mobile_robotic_chemist}. While the interpolated model is only approximate, close inspection suggested that it is broadly representative of the known real chemical space and sufficiently accurate to draw safe conclusions here. Our main goal is to test whether we can capture and inject experts' knowledge and intuition towards a better-informed and faster search. For a fair comparison, in place of TuRBO, LA-MCTS, and LA-MCTS+, we use the same DBO developed by ~\cite{c:Mobile_robotic_chemist} for experimental photocatalysis hydrogen production, capable of discretization, as a baseline.
\label{subsec:retrospective_knowledge}
\begin{figure}[t]
    \centering
    \includegraphics[width=0.7\columnwidth]{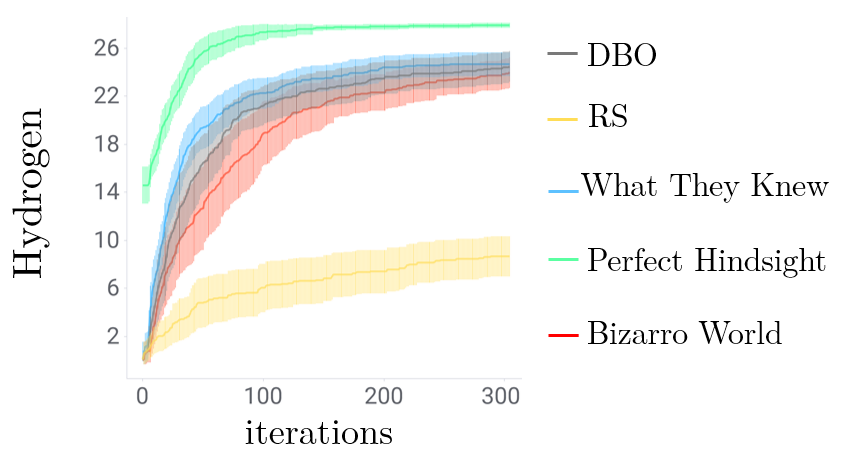}
    \caption{Retrospective application of hypotheses derived from \protect\cite{c:Mobile_robotic_chemist} using HypBO, compared to the no hypothesis run using DBO and RS. Shaded area is the standard deviation.}
    \label{fig:retrospective_hypotheses}
\end{figure}

\subsubsection{Retrospective Application of Knowledge} 
First, we used HypBO to fold in, retrospectively, knowledge of the underlying chemistry that was not captured in the \cite{c:Mobile_robotic_chemist} study to investigate whether injecting such hypotheses might improve performance. We explored three separate cases, outlined briefly below with more detailed explanations in the SM:
\begin{itemize}
    \item \textbf{What They Knew} In 2019, there was extra chemical knowledge available prior to the ~\cite{c:Mobile_robotic_chemist} study that could not have been injected using DBO; here, it is injected, retrospectively, using HypBO.
    \item \textbf{Perfect Hindsight} limits the search to within the optimal subspace based on \emph{post} \emph{facto} knowledge of the outcomes of all 1119 robotic experiments.
    \item \textbf{Bizarro World} purposefully focuses the search within the worst areas of the chemical space in all dimensions.
\end{itemize}

As shown in Figure \ref{fig:retrospective_hypotheses}, HypBO with `What They Knew' boosts performance somewhat in the early stages of the search, and overall it improves upon DBO, thus validating the benefits of considering expert hypotheses in real-world problems. For example, one can posit that any or all of the three dye components (MB, AR87, RB) might be beneficial but that high values would be counterproductive, based on chemical reasoning. We captured this in `What They Knew' by lowering the dyes' upper bounds ($\text{MB} \le 0.5\text{mL}$, $\text{AR87} \le 1\text{mL}$, $\text{RB} \le 0.5\text{mL}$). The somewhat modest boost given by `What They Knew' (Figure \ref{fig:retrospective_hypotheses}) can be explained by the partial knowledge available in 2019; indeed, some of `What They Knew' was, in fact, wrong. For example, as reported in \cite{c:Mobile_robotic_chemist}, all three dyes were strongly negative at all concentrations. We have not captured this post-experiment knowledge here; rather, `What They Knew' captures the knowledge that was available to this team in 2019, building on their initial formulation of hypotheses, prior to any robotic experiments.

Unsurprisingly, `Perfect Hindsight' leads to a much faster optimization. By contrast, although the artificially bad case of `Bizarro World' does lead to a slower search than DBO, the effects are greatly mitigated because HypBO can abandon unproductive hypotheses. 

\subsubsection{Searching the Chemistry Experiment Space with Mixed Hypotheses} \label{subsec:chem_experiment}
\begin{figure}[t]
    \centering
    \includegraphics[width=0.9\columnwidth]{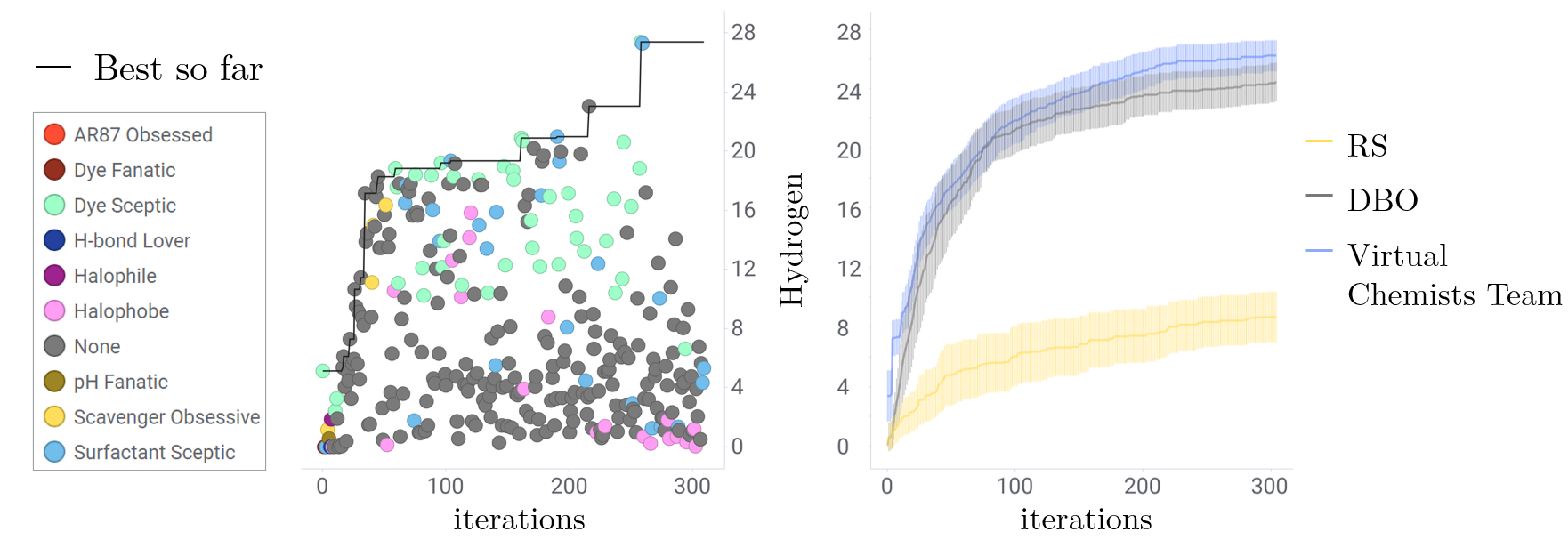}
    \caption{HypBO pruning the hypotheses and selecting seeds from the most promising ones to boost the optimization. Left: Scatter plot of the HypBO optimization with all nine virtual hypotheses; color denotes followed hypothesis, if any. Right: Best value obtained so far by HypBO using all nine hypotheses compared to DBO and RS.}
    \label{fig:fig_virtual_chemists}
\end{figure}
We test HypBO's ability to exploit good hypotheses and discard bad ones in a more realistic setting by creating a team of nine `virtual chemists', each with a virtual hypothesis based on plausible chemical reasoning detailed in the SM. The combined ``knowledge'' of this virtual team was then used to redo the simulated experiment for ~\cite{c:Mobile_robotic_chemist} in tandem with HypBO. The virtual team was designed to emulate the diverse and sometimes contradictory views of a real research team tackling a new problem. For example, some pairs of hypotheses (e.g., `Halophile' / `Halophobe') are in direct contradiction. Based on retrospective knowledge, `Dye Sceptic' and `Surfactant Sceptic' might be expected empirically to be the strongest hypotheses, while `Dye Fanatic' is probably the weakest. We applied all nine virtual hypotheses simultaneously and used the same oracle model described in the section above. For the purposes of these initial tests, all virtual hypotheses were considered of equal weighting. Figure \ref{fig:fig_virtual_chemists} illustrates the power of including human insights throughout the optimization. All hypotheses were selected initially, but as the optimization progressed, HypBO filtered out bad hypotheses and prioritized the most promising ones, improving the search compared to DBO and RS. While the least profitable hypotheses, such as `AR87 Obsessed', `Dye Fanatic', and `Halophobe', were deselected early on, HypBO does not discard them completely. For example, `Halophobe' was selected at times when its EI was greater than that of others that were available for evaluation. This captures the importance of re-evaluating hypotheses in the face of new data. Likewise, certain hypotheses are used while they are profitable and then discarded when they become delimiting; this can be observed for `Scavenger Obssessive', where some scavenger is indeed required, but not too much. The modest search improvement of the virtual chemist team over DBO (Figure 5, right) is somewhat arbitrary because we purposefully built this virtual team to be mediocre, with both ``good'' (informative) and ``bad'' (uniformative or misleading) hypotheses in near equal numbers.

\section{Conclusions}
\label{sec:conclusions}
To fully exploit the opportunities in laboratory robotics and automation, we need optimization methods that work in tandem with teams of human scientists. So far, BO has not fully leveraged the experience and hunches of experimenters. We harness that knowledge here by allowing them to inject their hypotheses about which parts of the input space will yield the best performance. We propose a BO variant, HypBO, that achieves this in a bi-level framework by recursively pruning and turning the hypotheses into seeds that augment sampling as a springboard for global optimization. HypBO stands out in its ability to concurrently handle and evaluate multiple expert-formulated hypotheses. It can also use weak hypotheses to converge faster than cases with no hypotheses and recover from poor ones. This highlights the power of human-computer interaction, re-imagining the role of humans in autonomous scientific discovery.
Future work includes initializing the hypotheses with weights based on the experimenter’s profile or confidence estimation. These weights might be particularly valuable in quick-starting hypothesis selection in large, diverse research teams, where expertise levels and domain specializations can vary quite widely. 

\section*{Supplementary Material}
\appendix
\section{Hypothesis Construction}
In this section, we describe the process to construct your hypothesis. A hypothesis involves using your domain knowledge and empirical observations as well as hunches to define your belief about the subspace of the search space containing the optimum or promising samples. To construct a hypothesis, you must translate that belief in terms of ranges and relationships. As described in the Methodology section of our work, we represent a hypothesis $\mathcal{H}$ as a system of manually designed constraints w.r.t. promising subspaces specified by an expert through a system of $p$ equations and $q$ inequalities describing an interval of confidence in the search space, see Equation \ref{eq:hypotheses_(in)equalities}.

\begin{example}[Maximizing the yield of a vegetable farm]
    The task is to maximize the yield in a vegetable farm. Let us assume there are three parameters for vegetable production with these ranges in this order:
    \begin{itemize}
        \item Watering: $ 0.5 \, \text{l/m}^2 \leq \text{Water} \leq 7 \, \text{l/m}^2 $
        \item Fertilization: $ 5 \, \text{g/m}^2 \leq \text{Fertilizer} \leq 85 \, \text{g/m}^2 $
        \item CO\textsubscript{2} Levels: $ 300 \, \text{ppm} \leq \text{CO\textsubscript{2}} \leq 1000 \, \text{ppm} $
    \end{itemize}
    
    The farmer might hypothesize that optimal conditions involve specific ranges for watering $ 1.5 \, \text{l/m}^2 \leq \text{Water} \leq 2.5 \, \text{l/m}^2 $ and fertilization  $ 20 \, \text{g/m}^2 \leq \text{Fertilizer} \leq 30 \, \text{g/m}^2 $, but has no idea about the optimal CO\textsubscript{2} level. The hypothesis schema accepted by HypBO, which is a writing in matrix form of the previous sentence, will be:
    
    \small
    \[
    \begin{bmatrix}
        -1  &   0   &   0   \\
        1   &   0   &   0   \\
        0   &   -1  &   0   \\
        0   &   1   &   0   \\ 
        0   &   0   &   -1  \\
        0   &   0   &   1   \\ 
    \end{bmatrix}
    \times
    x
    \leq
    \begin{bmatrix}
        -1.5    \\
        2.5     \\
        -20     \\
        30      \\
        -300    \\
        1000
    \end{bmatrix}
    \begin{array}{l}
        \rightarrow \text{means}\, 1.5 \, \text{l/m}^2 \leq \text{Water} \\
        \rightarrow \text{means}\, \text{Water} \leq 2.5 \, \text{l/m}^2 \\
        \rightarrow \text{means}\, 20 \, \text{g/m}^2 \leq \text{Fertilizer}\\
        \rightarrow \text{means}\, \text{Fertilizer} \leq 30 \, \text{g/m}^2\\
        \rightarrow \text{means}\, 300 \, \text{ppm} \leq \text{CO\textsubscript{2}}\\
        \rightarrow \text{means}\, \text{CO\textsubscript{2}} \leq 1000 \, \text{ppm}
    \end{array}
    \]

    These soft constraints form a hypothesis that HypBO will assess and focus its search within these specified ranges or in its vicinity if found promising, to find the optimal combination for the highest yield.
\end{example}

More examples of hypothesis creation can be found in Section \ref{appendix:retrospective_hypotheses} (Photocatalytic Hydrogen Production) of this document.

\subsection{Turning HypBO hypotheses into prior distribution for \texorpdfstring{$\mathbf{\pi BO}$}{piBO}}
To use $\pi$BO as a baseline in the Experiments section of our work, we converted our three hypotheses of different qualities (good, weak, and poor) into prior distributions. $\pi$BO can work with Gaussian priors in the following format ~\cite{c:piBO_prior_injection}:

\lstset{
    basicstyle=\ttfamily\footnotesize,
    breaklines=true,
    postbreak=\mbox{$\hookrightarrow$\space},
    language=Python,
    moredelim=**[is][\itshape\bfseries]{@}{@},
}

\begin{lstlisting}
prior_file = {
    ...
    "input_parameters": {
        "x0": {
            "parameter_type" : "real",
            "values" : [@x0_min_range, x0_max_range@],
            "prior" : "custom_gaussian",
            "custom_gaussian_prior_means": [@x0_mean@],
            "custom_gaussian_prior_stds": [@x0_std@]
        },
        ...
        "xn": {
            "parameter_type" : "real",
            "values" : [@xn_min_range, xn_max_range@],
            "prior" : "custom_gaussian",
            "custom_gaussian_prior_means": [@xn_mean@],
            "custom_gaussian_prior_stds": [@xn_std@]
        },
    },
    ...
}
\end{lstlisting}

Given that schema of prior injection and our schema of hypothesis, we converted each hypothesis into a $\pi$BO prior as described in Algorithm \ref{alg:hypbo_hypothesis_to_piBO_prior}:

\begin{algorithm}[tb]
    \caption{Converting HypBO hypothesis into a $\pi$BO Gaussian prior}
    \label{alg:hypbo_hypothesis_to_piBO_prior}
    \textbf{Input}: HypBO hypothesis $\mathcal{H}_j$, $\pi$BO's JSON prior file $prior\_file$.
    \textbf{Output}: {\raggedright The updated $\pi$BO prior file, \protect$prior\_file$}
    \begin{algorithmic}[1] 
        \STATE Get the HypBO hypothesis input parameter ranges by solving the system of equations and inequalities in Equation \ref{eq:hypotheses_(in)equalities};
        \FOR{each input parameter}
            \STATE The mean of the GP $\mu$ is the center of the input parameter range;
            \STATE The standard deviation of the GP $\sigma$ is 6\textsuperscript{th} of the range so that $99\%$ of the GP values fall within the range;
            \STATE Append the input parameter GP values to the $\pi$BO JSON file;
        \ENDFOR
        \RETURN The updated $\pi$BO prior file, $prior\_file$
    \end{algorithmic}
\end{algorithm}

\section{Ten-dimensional Model for Hydrogen Production}
\label{sec:HER_model}
\begin{figure*}[t]
    \centering
    \includegraphics[width=\textwidth]{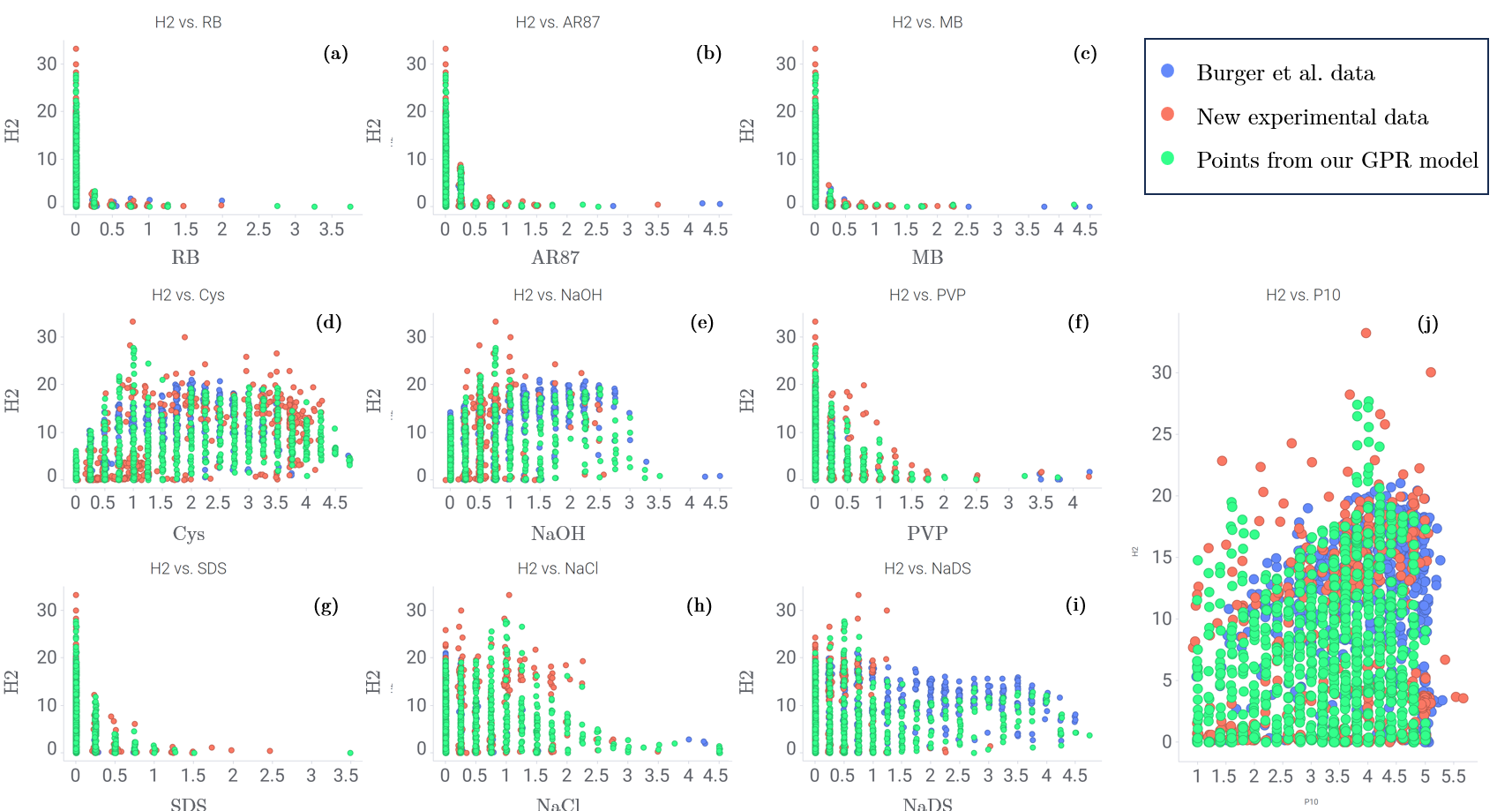}
    \caption{Plots showing the amount of hydrogen produced for 688 original experiments (blue points, \protect\cite{c:Mobile_robotic_chemist}) and for 431 new experiments, supplied by the same authors, conducted under the same conditions (red points). The green points (812 data points shown here) are derived from a model fitted against a combination of these two experimental datasets.}
    \label{fig:fig_hydrogen_amount}
\end{figure*}

In Section \ref{sec:her_experiment} of our work (Photocatalytic Hydrogen Production Optimization), we test HypBO against a real chemical problem. To do that, we constructed a ten-dimensional model describing hydrogen production for the experiments described in ~\cite{c:Mobile_robotic_chemist}. To do this, we augmented the original dataset of 688 experiments with a further 431 new experiments carried out under identical conditions and supplied by the authors of ~\citeauthor{c:Mobile_robotic_chemist} to create a total 'ground truth' dataset of 1119 experimental observations. The model was then built by fitting a Gaussian process regression (GPR) against the augmented dataset with a composite kernel consisting of Matern similarity, constant scaling and homoscedastic noise kernels. This hybrid kernel allows for variable smoothness and simulated experimental noise. Figure \ref{fig:fig_hydrogen_amount}, below, shows the experimental data (blue and red points) plotted along with 812 ``virtual'' data points derived from this model (green points) in all ten dimensions. These plots suggest that the model adequately captures the behavior described by the combined experimental dataset.
It should be noted that the new experiments found compositions that produce more hydrogen than any compositions reported in ~\cite{c:Mobile_robotic_chemist}. We believe that this is reasonable because the dimensionality of the search space (10 variables) is large compared to the original experiment run (688 experiments), and there was no guarantee of optimality in that search. Notably, the GPR model reproduces the strongly negative influence of the three dyes, RB, AR87 and MB, (Figure \ref{fig:fig_hydrogen_amount}a–c). It also appears that the new set of experiments (red points) might have discovered a new sub-space where NaCl contributes to the hydrogen production (Figure \ref{fig:fig_hydrogen_amount}h).

\section{Retrospective Hypotheses about the Photocatalytic Hydrogen Production}
\label{appendix:retrospective_hypotheses}
This section refers to the ``Retrospective Application of Knowledge'' subsection and describes the mathematical representation of the retrospective hypotheses we derived from the work of ~\citeauthor{c:Mobile_robotic_chemist} to use with HypBO. In that study, they detailed, in their description of the design of the experiment, some additional chemical knowledge they had about photocatalytic hydrogen production that was not captured by their discretized Bayesian optimizer (DBO). The optimizer could not account for this knowledge, which here constitutes the retrospective knowledge that we refer to in the paper as “What they knew”. Additionally, ~\citeauthor{c:Mobile_robotic_chemist} ’s chemistry experiment made 688 experimental observations that lasted eight days, extended here with a further 431 experiments supplied by the same team. By analyzing this augmented dataset of 1119 experiments, we derived two artificial items of knowledge. One is “Perfect Hindsight”, which represents the subspace of best samples found to maximize hydrogen production, while the other is “Bizarro World”, which represents the subspace of samples known to minimize hydrogen production.

\subsection{What They Knew}
To generate the hypothesis for the “\textbf{What they knew}” optimization test, a series of 10 sub-hypotheses was created based on mathematical constraints to capture the additional untapped chemistry knowledge of ~\citeauthor{c:Mobile_robotic_chemist}, or at least what they thought they knew, prior to the experiments that were carried out in ~\cite{c:Mobile_robotic_chemist}. These ten constraints, and the associated chemical rationale, are listed below. The hypotheses were constructed from a combination of facts stated in the ~\citeauthor{c:Mobile_robotic_chemist} paper and more general chemical reasoning that would be available to any typical chemist. For example, Burger et al. set up their experiments so that the total volume was always 5 mL:  as such, if 5 mL of any single component is added, then this does not leave any space for other components, and hence this might be too much.

\subsubsection{Hypothesis 1. \texorpdfstring{$\mathbf{P10} = 5\,\textnormal{mg}$}{P10 = 5 mg}}
\underline{Chemical rationale}: Generally, the amount of hydrogen produced in photocatalytic systems increases with the amount of semiconductor photocatalyst (in this case, P10).  Hence, in the total range of 1-5 mg that ~\citeauthor{c:Mobile_robotic_chemist} allowed, there was no obvious reason not to add the maximum amount of P10 since the hydrogen evolution rate—the search objective—was an absolute value that was not normalized to the mass of P10. This constraint (P10 = 5 mg) effectively removes one dimension from the search space.

\subsubsection{Hypothesis 2. \texorpdfstring{$\mathbf{Cys} = 1-4\,\textnormal{mL}$}{Cys =1-4 mL}}
\underline{Chemical rationale}: ~\citeauthor{c:Mobile_robotic_chemist} stated that some cysteine would be required; that is, $Cys > 0$, because this is a sacrificial hydrogen production reaction, but it was not known, prior to experiments, how much cysteine would be optimal. However, because they constrained the maximum volume of the mixture of compounds to be 5 mL, then it might be undesirable to set, say, $Cys > 4 mL$ since this would not then allow enough volume for adding other components. Using this simple reasoning, a range of $Cys = 1-4 mL$ can be hypothesized, rather than the range of $Cys = 0-5 mL$ that was used in the original experiments in ~\cite{c:Mobile_robotic_chemist}. 

\subsubsection{Hypothesis 3. \texorpdfstring{$\mathbf{MB} < 0.5\,\textnormal{mL}$}{MB < 0.5 mL}}
\underline{Chemical rationale}: ~\citeauthor{c:Mobile_robotic_chemist} expressed that adding dyes to the reaction was based on earlier observations that dyes could sensitize related photocatalysts, thus increasing hydrogen production ~\cite{c:photocatalytic_hydrogen_ev_from_water}. However, as mentioned in ~\cite{c:Mobile_robotic_chemist}, it was not known before experiments whether these three dyes (MB, AR87, RB) would be positive or negative in the case of P10. Still, one might reasonably expect that very high dye concentrations (e.g., $> 4 mL$) could (a) absorb all of the light, thus lowering the hydrogen evolution rate, and (b) not leave enough volume for other components, as argued for Cys, above. Given the high light absorption coefficients of these three dyes, we conjecture that ‘a little’ might be enough; as such, a constraint of $MB < 0.5 mL$ was formulated.

\subsubsection{Hypothesis 4. \texorpdfstring{$\mathbf{RB} < 0.5\,\textnormal{mL}$}{RB < 0.5 mL}}
\underline{Chemical rationale}: Exactly as for MB (Hypothesis 3).

\subsubsection{Hypothesis 5. \texorpdfstring{$\mathbf{AR87} < 1\,\textnormal{mL}$}{AR87 < 1 mL}}
\underline{Chemical rationale}: As above for the dyes MB and AR87 except that here, the upper range, 1 mL, is larger because there was pre-existing evidence in ~\cite{c:photocatalytic_hydrogen_ev_from_water} that this dye (called Eosin Y in that paper) could sensitize a related photocatalyst. As such, this dye might have been given a higher upper bound prior to experiments even though, in reality, ~\citeauthor{c:Mobile_robotic_chemist} found it to be a negative component in the reaction mixture.

\subsubsection{Hypothesis 6. \texorpdfstring{$\mathbf{NaOH} < 3\,\textnormal{mL}$}{NaOH < 3 mL}}
\underline{Chemical rationale}: ~\citeauthor{c:Mobile_robotic_chemist}'s rationale for including NaOH was that solution pH might influence hydrogen production, but it was not known whether this would be positive or negative. A chemist might also have a potential concern that high concentrations of NaOH might degrade other components in the mixture, although this was not discussed by Burger et al. Also, as for Cys (Hypothesis 2), there is an argument for constraining the volume of NaOH to allow free volume for adding other components. These combined arguments led us to an estimated constraint of $NaOH < 3 mL$, allowing (unlike Hypothesis 2) for a value of NaOH = 0 since from our reading of Burger et al. it was not obvious, \textit{a priori} before experiment, that NaOH addition would have any positive effect.

\subsubsection{Hypothesis 7. \texorpdfstring{$\mathbf{NaCl} < 3\,\textnormal{mL}$}{NaCl < 3 mL}}
\underline{Chemical rationale}: There was no \textit{a priori} evidence that NaCl would be either positive or negative before experiments; as such, the constraint of 3 mL is to allow sufficient free volume for other components, exactly as for Hypothesis 6 above.

\subsubsection{Hypothesis 8. \texorpdfstring{$\mathbf{SDS} < 1\,\textnormal{mL}$}{SDS < 1 mL}}
\underline{Chemical rationale}: It was argued in ~\cite{c:Mobile_robotic_chemist} that the primary rationale for adding surfactants was that it could aid in dispersing the photocatalysis, P10, in water. There was no evidence for this prior to the experiment. Still, given the highly surface-active nature of SDS, we might have hypothesized that a little surfactant could be enough, as argued for the three dyes (Hypotheses 3–5), hence $SDS < 1 mL$.

\subsubsection{Hypothesis 9. \texorpdfstring{$\mathbf{PVP} < 2\,\textnormal{mL}$}{PVP < 2 mL}}
\underline{Chemical rationale}: PVP is also a surfactant, and hence the rationale is the same as for SDS (Hypothesis 7) – the larger range allowed here (maximum 2 mL instead of 1 mL) is because PVP is less surface-active than SDS at a given concentration. Hence, a higher volume might be required to be effective.

\subsubsection{Hypothesis 10. \texorpdfstring{$\mathbf{NaDS} < 4\,\textnormal{mL}$}{NaDS < 4 mL}}
\underline{Chemical rationale}: As for NaOH (Hypothesis 6) – no prior information was available for the effect of this component, so the constraint arises from the need to allow some residual volume for other components, e.g., for CyS, which should be non-zero (see Hypothesis 2).

\subsection{Perfect Hindsight}
This hypothesis limits the search to within the optimal subspace based on \emph{post} \emph{facto} knowledge of the outcomes of all 1119 robotic experiments (see Figure \ref{fig:fig_hydrogen_amount}). It is described by the following ten constraints:
\begin{enumerate}
    \item $P10 \ge 3.5\ \text{mg}$
    \item $1\ \text{mL} \le Cys \le 3.5\ \text{mL}$
    \item $0.5\ \text{mL} \le NaOH \le 2\ \text{mL}$
    \item $0\ \text{mL} \le NaDS \le 1.5\ \text{mL}$
    \item $2\ \text{mL} \le Cys + NaOH + NaDS \le 4.5\ \text{mL}$
    \item $1\ \text{mL} \le NaCl + NaDS + NaOH \le 2.75\ \text{mL}$
    \item $MB = 0\ \text{mL}$
    \item $AR87 = 0\ \text{mL}$
    \item $RB = 0\ \text{mL}$
    \item $NaCl < 2.5\ \text{mL}$
    \item $SDS = 0\ \text{mL}$
    \item $PVP = 0\ \text{mL}$
\end{enumerate}

\subsection{Bizarro World}
This hypothesis limits the search to within the subspace of samples known to minimize the hydrogen production, based on \emph{post} \emph{facto} knowledge of the outcomes of all 1119 robotic experiments. It is described by the following ten constraints:
\begin{enumerate}
    \item $P10 = 1\ \text{mg}$
    \item $Cys = 0\ \text{mL}$
    \item $MB > 0.5\ \text{mL}$
    \item $AR87 > 0.5\ \text{mL}$
    \item $RB > 0.5\ \text{mL}$
    \item $NaOH = 0\ \text{mL}$
    \item $NaCl > 0.5\ \text{mL}$
    \item $SDS > 0.5\ \text{mL}$
    \item $PVP > 0.5\ \text{mL}$
    \item $NaDS = 0\ \text{mL}$
\end{enumerate}

\begin{table*}[t]
    \centering
    \small
    \begin{tabular}{ l l l l l l l l l}
        \hline
        HypBO Variant   & HypBO$^{10}_{1}$  & HypBO$^{10}_{2}$  & HypBO$^{10}_{3}$  & ...   & HypBO$^{10}_{10}$ & HypBO$^{1}_{10}$  & HypBO$^{5}_{2}$ \\ \hline
        $u$       & 10                & 10                & 10                & ...   & 10                & 1                 & 5 \\
        $l$       & 1                 & 2                 & 3                 & ...   & 10                & 10                & 2 \\
        \hline
    \end{tabular}
    \caption{Details of the HypBO variants used in the ablation study.}
    \label{t:hypbo_variants}
\end{table*}

\section{Virtual Chemists} \label{appendix:virtual_chemists}
This section refers to the subsection ``Searching the Chemistry Experiment Space with Mixed Hypotheses'', in which we test HypBO's ability to exploit good hypotheses and discard bad ones in a more realistic setting. This test mimics an experiment designed by a research team that holds a range of different opinions. To do this, we created nine ‘virtual chemists’ who hold hypotheses based on plausible chemical reasoning, which is outlined below. We purposefully designed this virtual team to be in conflict, with ‘good’ and ‘bad’ hypotheses in near equal numbers.  As such, this is a deliberately mediocre virtual team: it was built to test the ability of HypBO to sort good and bad hypotheses, rather than to create a performance boost in the optimization, although as shown in the main text, it does in fact give a small improvement to the search.

This virtual chemist team reveals how each virtual chemist brings unique perspectives and strategies to the optimization process. It is also important to note a parallel with ~\cite{c:coexbo}. This work highlights the importance of integrating varied human expert perspectives into Bayesian optimization, and is a concept that aligns with our approach utilizing virtual chemists.

\subsection{Virtual Chemist \#1: “Dye Sceptic”} This chemist does not believe that dyes will have a positive effect on the reaction but has no other opinions about the value of other components in the mixture (that is, all other variables are allowed to range between the original low–high values when this hypothesis is applied). This hypothesis is expressed by the following constraints:
\begin{enumerate}
    \item $MB = 0\ \text{mL}$
    \item $AR87 = 0\ \text{mL}$
    \item $RB = 0\ \text{mL}$
\end{enumerate}

\subsection{Virtual Chemist \#2: “Dye Fanatic”} This chemist believes that dyes will have a positive effect and that a high dye concentration is required to achieve this; for example, to push the chemical equilibrium to achieve surface dye absorption on the photocatalyst. However, this chemist does not know which dye is best (all three are equally likely). Hence the \textit{total} dye concentration is constrained to be greater than 3 mL; again, this chemist has no opinions about other components in the reaction. This hypothesis is expressed by the following constraints:
\begin{enumerate}
    \item $MB + AR87 + RB > 3\ \text{mL}$
\end{enumerate} 

\subsection{Virtual Chemist \#3: “AR87 Obsessed”} Like Virtual Chemist \#2, this chemist believes that dyes will have a positive effect but is convinced moreover that a specific dye, AR87, is optimal having read the earlier report that this dye was successful with other related photocatalysts ~\cite{c:photocatalytic_hydrogen_ev_from_water}. Virtual Chemist \#3 believes that the other two dyes are likely to be less effective than AR87. As such, the recommendation of Virtual Chemist \#3 is to ‘play in the dye space’ but with a strong emphasis on AR87, as expressed by:
\begin{enumerate}
    \item $AR87 > 3\ \text{mL}$
    \item $MB < 0.5\ \text{mL}$
    \item $RB < 0.5\ \text{mL}$
\end{enumerate}

\subsection{Virtual Chemist \#4: “Surfactant Sceptic”} This chemist believes that surfactants will be bad for the reaction but has no other opinions about any other components, as expressed by:
\begin{enumerate}
    \item $SDS = 0\ \text{mL}$
    \item $PVP = 0\ \text{mL}$
\end{enumerate}

\subsection{Virtual Chemist \#5: “Scavenger Obsessive”} This chemist is convinced that a very high concentration of the scavenger, Cys, is needed to achieve high levels of hydrogen production, as expressed by:
\begin{enumerate}
    \item $Cys > 4\ \text{mL}$
\end{enumerate}

\subsection{Virtual Chemist \#6: “pH Fanatic”} This chemist believes high pH is needed to boost hydrogen production. Both NaOH and NaDS are bases and will increase the pH, and this chemist considers both to be interchangeable, leading to the hypothesis:
\begin{enumerate}
    \item $NaOH + NaDS > 3.5\ \text{mL}$
\end{enumerate}

\subsection{Virtual Chemist \#7: “H-bond Lover”} This chemist believes that H-bonding with NaDS will lead to increased hydrogen production, based on reading earlier publications made similar hypotheses about NaDS, e.g., ~\cite{c:H_bonding_effect}, leading to the hypothesis: 
\begin{enumerate}
    \item $NaDS > 3.5\ \text{mL}$
\end{enumerate}

\subsection{Virtual Chemist \#8: “Halophile”} This chemist believes that high ionic strength is crucial for the success of the reaction–that is, a high concentration of salt; NaOH, NaDS and NaCl are all salts, leading to the hypothesis:
\begin{enumerate}
    \item $NaOH + NaDS + NaCl > 3.5\ \text{mL}$
\end{enumerate}

\subsection{Virtual Chemist \#9: “Halophobe”} In contrast to Virtual Chemist \#8, this chemist believes that high NaCl concentration is bad for the reaction because they half-remember seeing a conference presentation that claimed that NaCl addition could lead to chlorine gas production, rather than hydrogen production. This leads to the hypothesis:
\begin{enumerate}
    \item $NaCl = 0\ \text{mL}$
\end{enumerate}

\section{Ablation Studies}
Here, we study HypBO's sensitivity to its hyper-parameters. In Section \ref{sec:u_l_ablation_study}, we focus on the impact of the choice of $u_{max}$ and $l_{max}$ before delving into $\gamma$'s effect in Section \ref{sec:gamma_ablation_study}.

\subsection{\texorpdfstring{$\mathbf{l_{max}}$}{} vs \texorpdfstring{$\mathbf{u_{max}}$}{} Ablation Study}
\label{sec:u_l_ablation_study}
We conducted an ablation study to analyze how sensitive HypBO is to the lower and upper failure thresholds $l_{max}$ and $u_{max}$ in the Convergence Criteria section \ref{sec:convergence} of our research. Our goal was to find if there was an optimal balance of upper and lower level failure thresholds. To achieve this, we repeated the optimization experiments on the synthetic functions with a single hypothesis (Poor, Weak, and Good) and mixed hypotheses (Poor \& Good, Poor \& Weak, and Weak \& Good) as described in Section 4.2 (Synthetic Functions) of our work with different combinations of $l_{max}$ and $u_{max}$. We fixed $u_{max}$ to 10 and incrementally increased $l_{max}$ from 1 to 10, getting ten variants of HypBO spanning from a slight to an equal focus on the hypothesis regions. We extended these variants with two other variants:
\begin{itemize}
    \item The strong focus case on the hypothesis regions with $u_{max}=1$ and $l_{max}=10$.
    \item The default case in our work where $u_{max}=5$ and $l_{max}=1$.
\end{itemize}

For notation simplicity, we will refer to $u_{max}$ as $u$, and $l_{max}$ as $l$. We ran the experiments against those twelve variants as reported in Table \ref{t:hypbo_variants}.  The maximum number of iterations was limited to 100, and the result of 50 repeated trials was reported as the mean value.

\begin{figure}[t]
    \centering
    \includegraphics[width=\columnwidth]{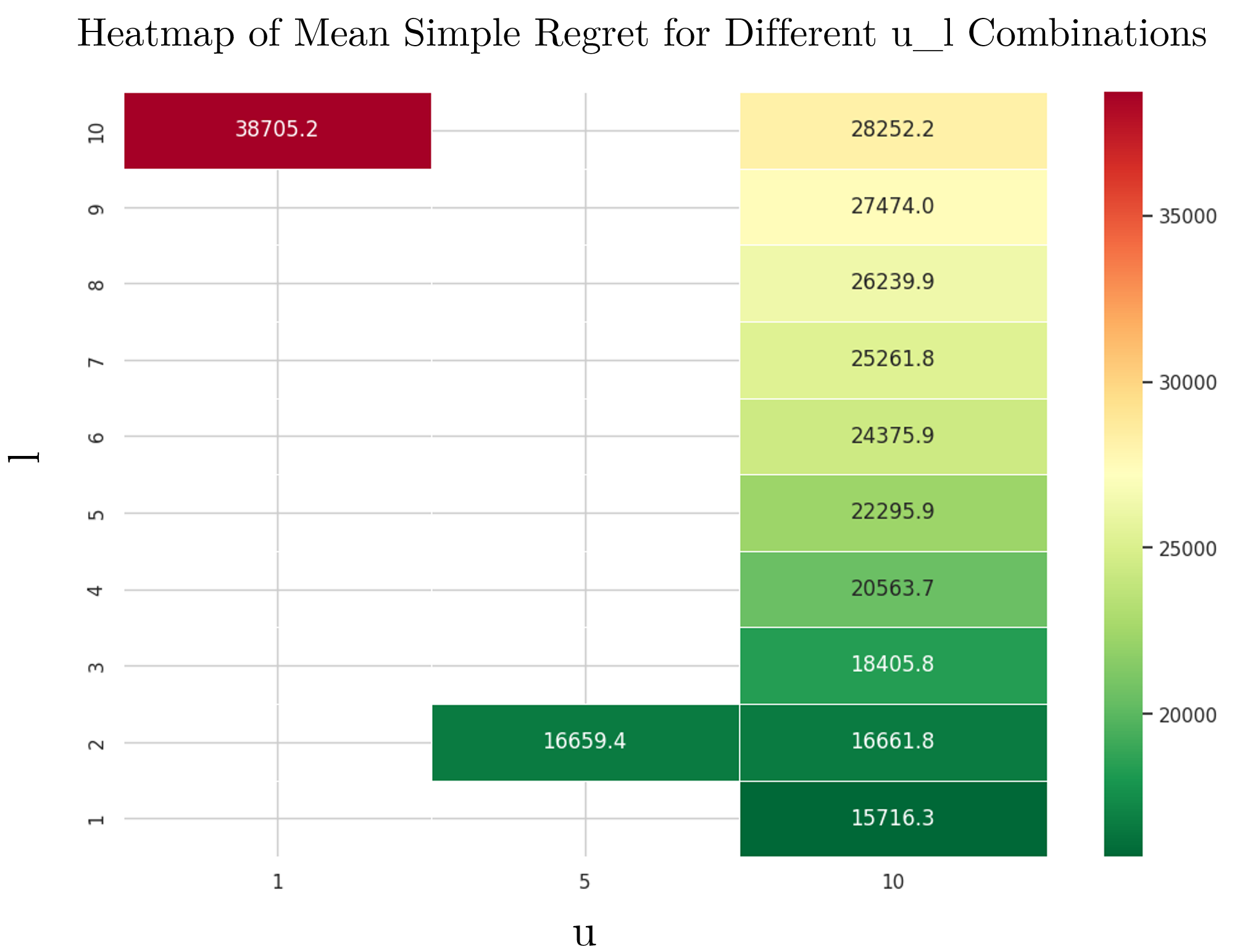}
    \caption{Heatmap of the mean simple regret for different combinations of \protect$u$ and \protect$l$}
    \label{fig:fig_ablation_study_heatmap_regret_u_l}
\end{figure}

\begin{figure*}[t]
    \centering
    \includegraphics[width=\textwidth]{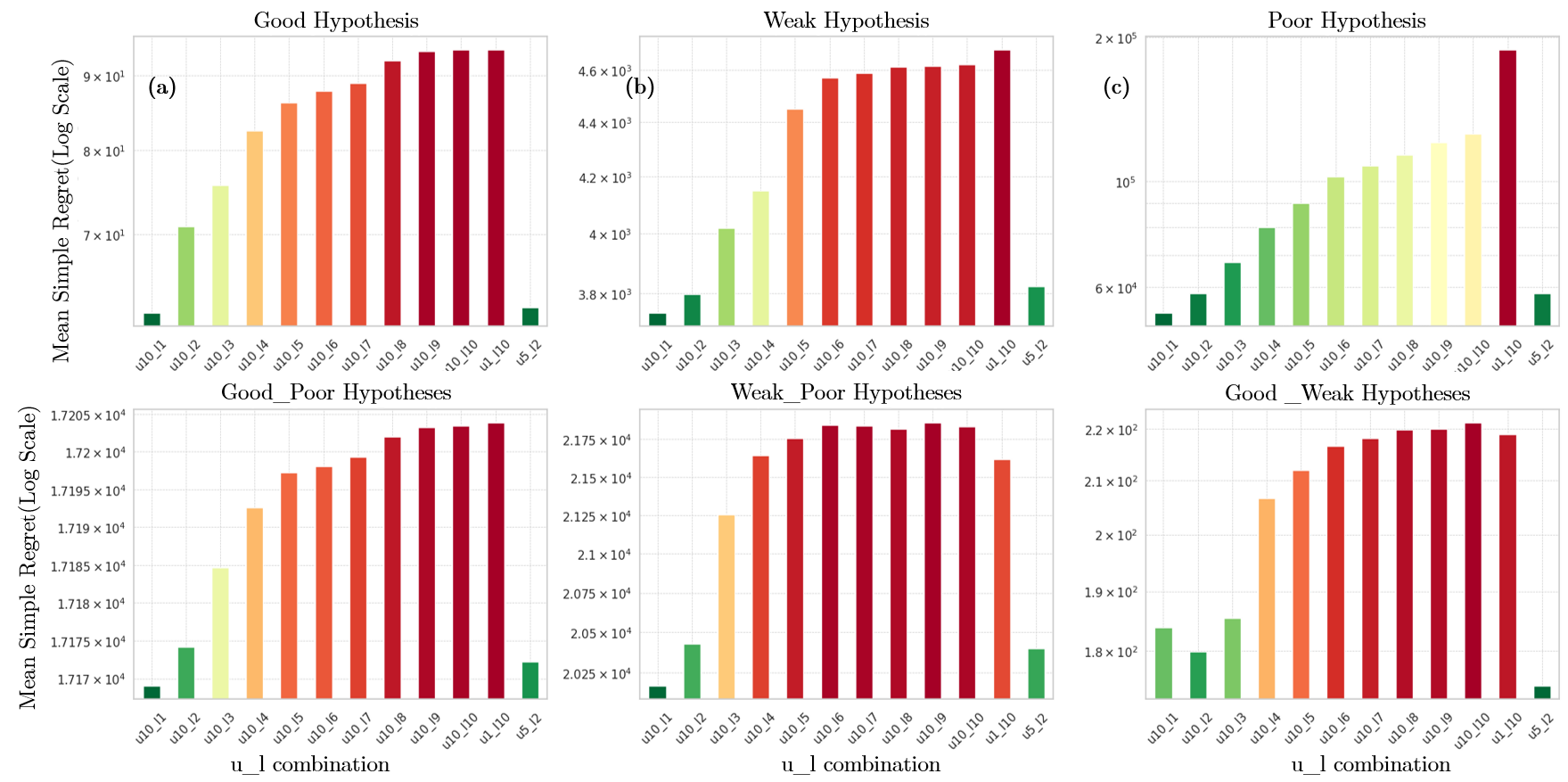}
    \caption{Evolution of the mean simple regret across all synthetic functions of different hypothesis mixtures for each combination of \protect$u$ and \protect$l$}
    \label{fig:fig_ablation_study_bar_plots}
\end{figure*}

\begin{figure}[t]
    \centering
    \includegraphics[width=\columnwidth]{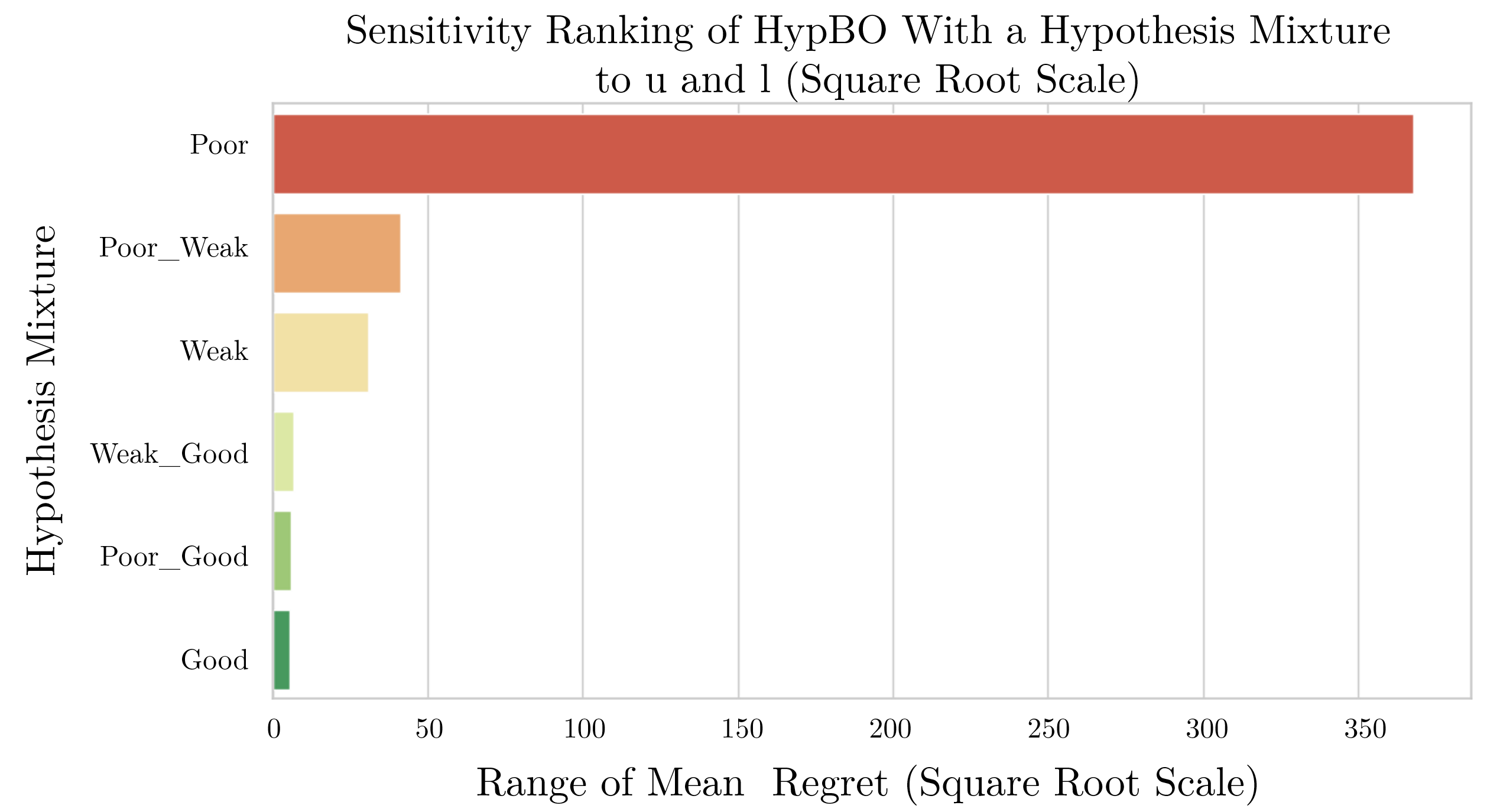}
    \caption{Sensitivity ranking of HypBO with different hypothesis scenarios to \protect$u$ and \protect$l$ Longer bars indicate greater sensitivity of HypBO under that scenario, i.e, its regret is more impacted by changes in \protect$u$ and \protect$l$.}
    \label{fig:fig_ablation_study_sensitivity_hypothesis}
\end{figure}

Figure \ref{fig:fig_ablation_study_heatmap_regret_u_l} and \ref{fig:fig_ablation_study_bar_plots} show that there seems to be a trend with the regret increasing with $l$ and decreasing with $u$, meaning a strong focus on the hypothesis regions might be detrimental. They also show that $u10\_l1$ seems to be the optimal combination of $u$ and $l$ that minimizes regret. These findings are confirmed with a regression analysis under the assumption that the relation between $u\_l$ and the regret is linear.

Concerning the sensitivity of HypBO, under each hypothesis scenario, to $u$ and $l$, Figure \ref{fig:fig_ablation_study_sensitivity_hypothesis} reveals that the sensitivity decreases with better hypothesis case qualities. Overall, all scenarios are robust to changes in $u$ and $l$ except for the pure poor hypothesis case.

\subsubsection{Considerations of using \texorpdfstring{$u5\_l2$}{} as the default}
\begin{figure}[t]
    \centering
    \includegraphics[width=\columnwidth]{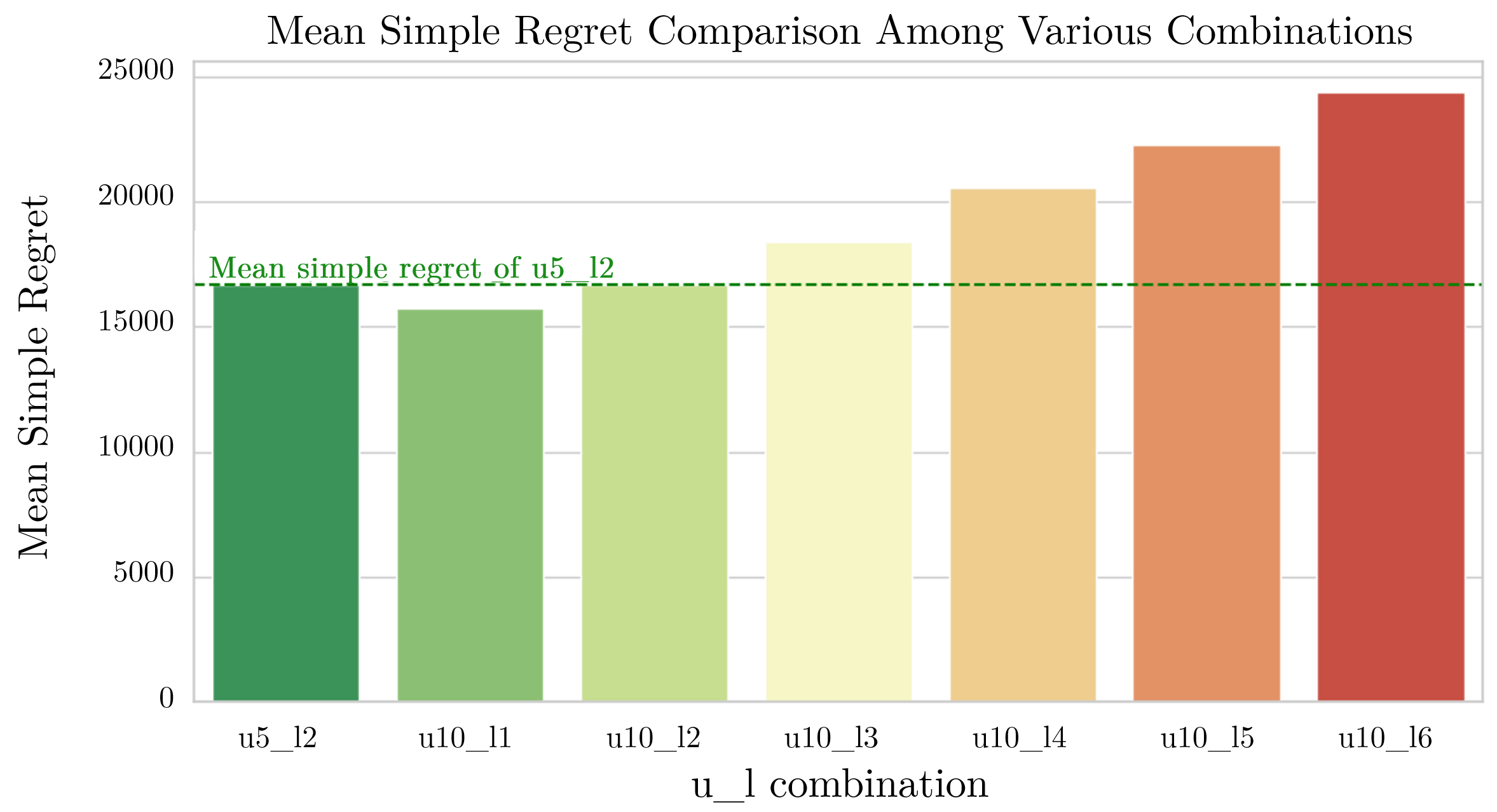}
    \caption{Visual comparison of the mean regrets of various parameter combinations, including \protect$u10\_l1$ and \protect$u5\_l2$. The green dashed line represents \protect$u5\_l2$ 's mean regret, which allows for a direct comparison with other combinations}
    \label{fig:fig_ablation_study_u10_1_vs_u5_l2}
\end{figure}

\begin{figure*}[t]
    \centering
    \includegraphics[width=\textwidth]{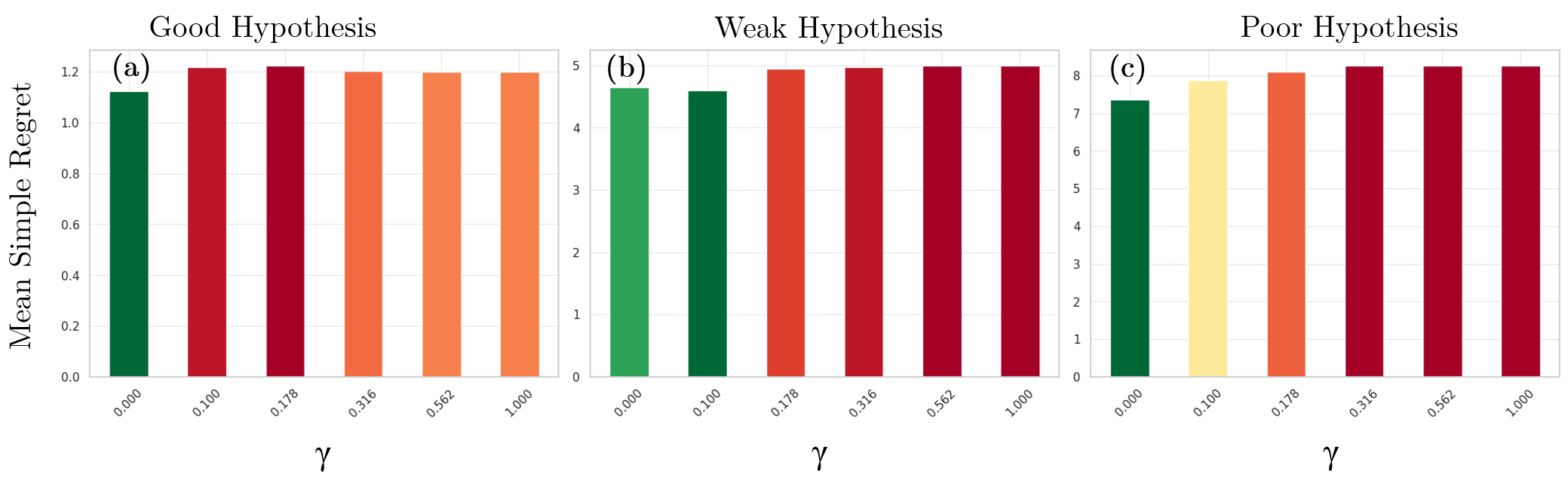}
    \caption{Evolution of the mean simple regret of different hypotheses with \protect$\gamma$}
    \label{fig:fig_gamma_ablation_study_bar_plots}
\end{figure*}

\begin{figure}[t]
    \centering
    \includegraphics[width=\columnwidth]{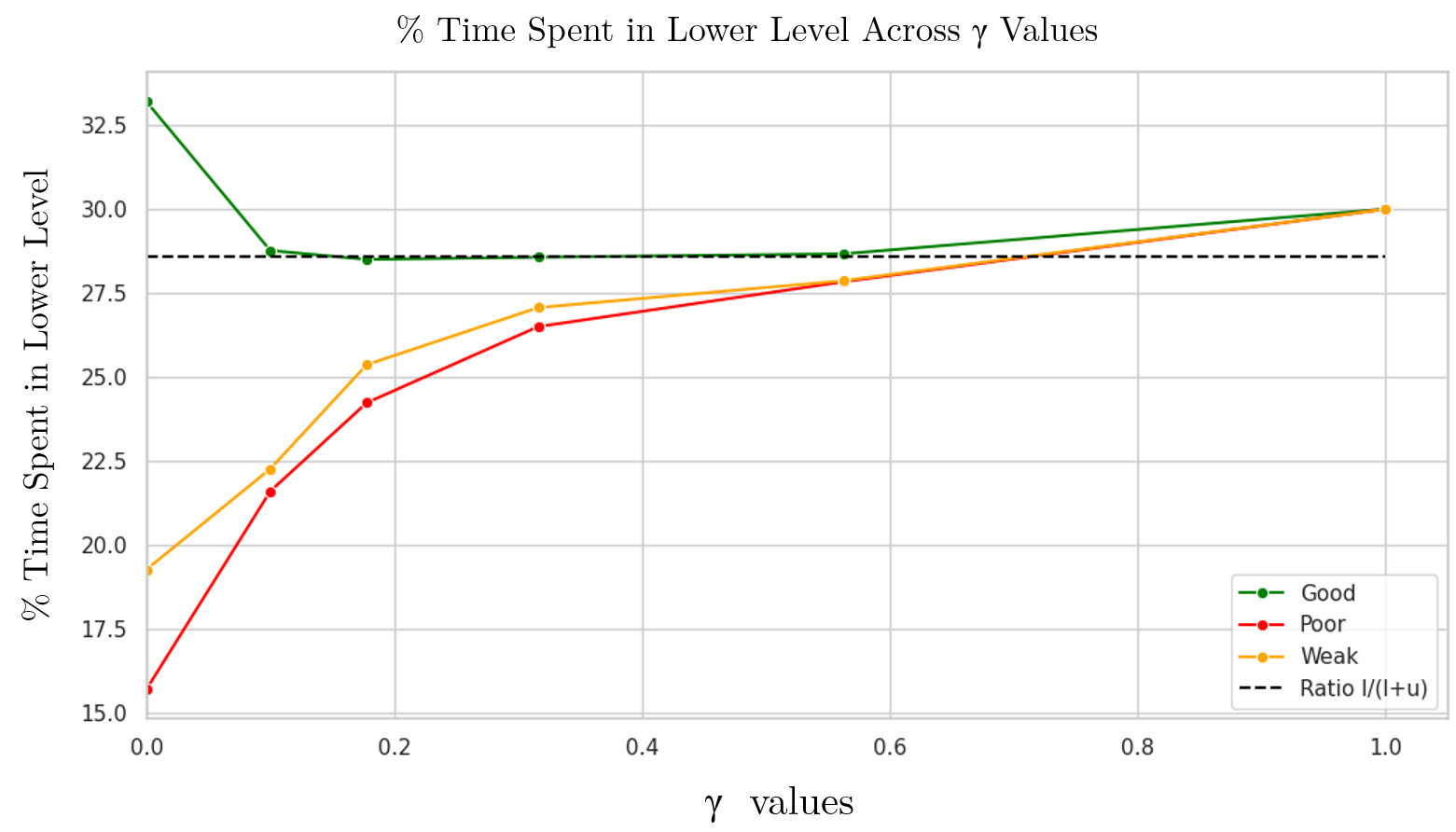}
    \caption{Percentage of the number of iterations (time) spent in the lower level across \protect$\gamma$. The black dashed line represents the ratio \protect$\frac{l}{l+u}$}
    \label{fig:fig_gamma_ablation_study_time_spent}
\end{figure}

From the previous results, $u10\_l1$ is the optimal choice. However, they are practical and other considerations to take into account for the choice of $u$ and $l$, which might motivate choosing a different $u\_l$ combination. In our work, we chose $u5\_l2$. $u10\_l1$'s mean regret is just 5\% less than $u5\_l2$'s mean regret. Figure \ref{fig:fig_ablation_study_u10_1_vs_u5_l2} shows that $u5\_l2$ performs relatively well compared to several other $u\_l$ combinations but is outperformed by $u10\_l1$. We extended this comparison with the following considerations:
\begin{itemize}
    \item \textbf{Balanced Approach}: The combination $u5\_l2$ offers a relatively balanced trade-off between exploring the entire input search ($u=5$) and focusing on the hypothesis subspace ($l=2$). This balance can be advantageous in various scenarios where neither extreme exploration nor focused exploitation is desired.
    \item \textbf{Robustness}: Because $u5\_l2$ shows consistent performance across different functions and hypothesis mixtures, it could be a more robust choice, providing reasonably good results in a wide range of scenarios.
    \item \textbf{Practicality}: In black-box scientific experiments, we assume that good hypotheses with narrow ranges are rare. Most experimenters would provide weak ones and possibly a combination of weak and good hypotheses, as we saw with the ``What They Knew'' scenario of Section \ref{appendix:retrospective_hypotheses}. In those scenarios, Figure \ref{fig:fig_ablation_study_bar_plots} (c and f) shows that $u5\_l2$ can outperform $u10\_l1$.
\end{itemize}

In summary, although $u5\_l2$ may not be the optimal choice for minimizing regret, it offers a well-balanced and potentially more robust and practical option as a default setting. However, it is essential to further validate this recommendation by taking into account the specific characteristics and requirements of the optimization problem being tackled with HypBO.

\subsection{\texorpdfstring{$\mathbf{\gamma}$}{} Ablation Study}
\label{sec:gamma_ablation_study}
$\gamma$ sets the percentage of improvement over the best $y$ found so far, which is considered ``significant" for the optimization process. In this section, we present an analysis of the sensitivity of HypBO to the growth rate $\gamma$ in the Convergence Criteria section \ref{sec:convergence} of our research. We conducted optimization experiments on the synthetic functions Branin\textsubscript{d2}, Sphere\textsubscript{d2} and Ackley\textsubscript{d5}, using a single hypothesis (Poor, Weak, and Good), five initial samples for a budget of 50 iterations, and 20 trials.

Figure \ref{fig:fig_gamma_ablation_study_bar_plots} demonstrates that HypBO is relatively robust to changes in $\gamma$ with some slight trends. With a Good hypothesis, HypBO's mean regret slightly decreases  with an increase in $\gamma$, while with a Weak / Poor hypothesis, its mean regret slightly increases with $\gamma$. This observation correlates with the time HypBO spends in the lower level, i.e., hypothesis region, increasing with $\gamma$ as shown in Figure \ref{fig:fig_gamma_ablation_study_time_spent}. As increasing $\gamma$ makes the criteria for progress more strict, HypBO oscillates more between the upper and the lower levels. Consequently, this brings the percentage of the time spent in the lower level closer to the ratio $\frac{l}{l+u}$. It is important to note that because the Good hypothesis contains the optimum, the percentage of time spent in its region is somewhat arbitrary.

In summary, the performance of HypBO remains consistent across $\gamma$. However, it is worth noting that the performance shows a marginal decline as $\gamma$ increases under a Weak or Poor hypothesis,  while under a Good hypothesis, HypBO is quite robust to changes in $\gamma$. As a result, we recommend a maximum $\gamma$ value of 10\%.

\section{Reproducibility}
This section details the steps to reproduce the experiments and results in \textit{HypBO: Expert Hypotheses to Guide Bayesian Search in Material Discovery}.

\subsection{Experimental Setup}
Here, we describe our experiments' specific configurations, parameters, and environments. This includes the detailed settings of the baselines and the software specifications and versions.

\subsubsection{Baselines}
\begin{itemize}
    \item \textbf{Random Search (RS)} Random search under uniform distribution over the search space.
    \item \textbf{Trust Region Bayesian Optimization (TuRBO)} We set the number of trust regions $m=1$, the success threshold $\tau_{succ} = 3$, the failure threshold $\tau_{fail} = [d/q]$ where $d$ is the number of dimensions and $q$ is the batch. The batch size is $q=1$. The base side length $L_{init} = 0.8$, its minimum $L_{min} = 2^{-7}$ and maximum $L_{max} = 1.6$.
    \item \textbf{Latent Action Monte Carlo Tree Search (LA-MCTS)} We set LA-MCTS' hyperparameters as follows: $C_p=1$, $\theta=10$, gamma type is auto, SVM boundary kernel is RBF, and the solver is BO with a Matern kernel.
    \item \textbf{LA-MCTS with hypothesis-based initial design (LA-MCTS+)} We modified the previous baseline to initialize it exclusively within the hypothesis subspaces before the regular search in the entire search space. Its hyperparameters' values are the same as those of LA-MCTS.
    \item $\bm{\pi}$\textbf{BO} We kept the default value of $\beta = i_{max}/10$, where $i_{max}$ is the total number of iterations ~\cite{c:piBO}.
\end{itemize}

\subsubsection{Software Specifications}
To ensure consistency in software environments, each baseline environment is a Docker container whose Dockerfile is saved in the baseline's folder in the code repository. Docker is a tool that simplifies application creation, deployment, and running using containers. Unlike Python environments, these containers encapsulate an application’s entire runtime environment, including system libraries and settings. Because Docker containers are not affected by variations in host system configurations, Docker is well-suited for replicating complex applications' environments, which reduces "it works on my machine" problems, a level of consistency that Python virtual environments can't provide on their own.

\subsection{Data Accessibility}
All datasets used in this study, including the dataset used to train the GPR to replicate the Photocatalytic Hydrogen Production Optimization experiment in ~\cite{c:Mobile_robotic_chemist}, as well as that replica and the synthetic functions experiments' results, and ablation studies, are made publicly available at the Anonymous Github repository, HypBO, in the data folder. We provide the link to the HypBO repository https://anonymous.4open.science/r/HypBO/, ensuring that researchers can access and utilize the exact data for replication.

\subsection{Source Code}
To ensure that our results can be replicated exactly, we have made available the complete source code of HypBO, including scripts for data preprocessing, model training, and result analysis. This source code also includes the source codes of the baseline methods LA-MCTS (https://github.com/facebookresearch/LaMCTS/), $\pi BO$ (https://github.com/luinardi/hypermapper/wiki/Prior-Injection), and TuRBO (https://github.com/uber-research/TuRBO) cloned from their respective repositories. HypBO source code has been anonymized and can be accessed via the Anonymous Github https://anonymous.4open.science/r/HypBO/. Furthermore, we have provided clear documentation on how to run the experiments and reproduce the results.

\section*{Acknowledgements}
The authors acknowledge financial support from the Leverhulme Trust via the Leverhulme Research Centre for Functional Materials Design. AIC thanks the Royal Society for a Research Professorship (RSRP\textbackslash S2\textbackslash 232003).

\bibliographystyle{named}
\bibliography{ijcai24}

\end{document}